\documentclass[]{fairmeta}
\usepackage{pifont}

\usepackage{amsmath,amsfonts,bm}

\def\eqref#1{equation~\ref{#1}}
\def\1{\bm{1}}

\DeclareMathAlphabet{\mathsfit}{\encodingdefault}{\sfdefault}{m}{sl}
\SetMathAlphabet{\mathsfit}{bold}{\encodingdefault}{\sfdefault}{bx}{n}

\usepackage{bm}
\usepackage{bbm}
\usepackage{multirow}
\usepackage{pifont}
\usepackage{algorithm}
\usepackage{wrapfig}
\usepackage{colortbl}
\usepackage{booktabs}

\usepackage{amsmath}
\usepackage{amssymb}
\usepackage{mathtools}
\usepackage{amsthm}
\usepackage[most]{tcolorbox}
\usepackage{graphicx}
\usepackage{hyperref,xcolor}
\usepackage{enumitem}
\usepackage{fontawesome5}
\theoremstyle{plain}

\theoremstyle{definition}

\theoremstyle{remark}

\usepackage[dvipsnames]{xcolor}

\definecolor{midnightgreen}{rgb}{0.0, 0.29, 0.33}
\definecolor{deepgreen}{HTML}{055c29}
\definecolor{deeppurple}{HTML}{7030a0}
\definecolor{deepblue}{HTML}{171d91}
\definecolor{brown}{HTML}{843c0c}
\definecolor{shadered}{HTML}{ffe5e5}
\definecolor{shadegreen}{HTML}{e5f7ed}
\definecolor{msftBlack}{RGB}{0,0,0}
\definecolor{lightred}{RGB}{255,163,163}
\definecolor{deepred}{RGB}{153,0,0}
\definecolor{softblue}{RGB}{30, 90, 160}

\definecolor{barblue}{RGB}{90,120,180}
\definecolor{barorange}{RGB}{225,124,5}

\definecolor{DeltaBg}{HTML}{D4F2D7} % green
\definecolor{SearchBg}{HTML}{C2E6F5} % blue
\definecolor{AgenticBg}{HTML}{F5C2CC} % red
\definecolor{MathBg}{HTML}{E6D4F2} % purple
\definecolor{ScienceBg}{HTML}{FBE0BC} % yellow

\definecolor{mygreen}{RGB}{229, 245, 224}
\definecolor{mygray}{RGB}{242, 242, 242} 
\definecolor{myred}{RGB}{248, 230, 234}      % Light red-gray background

\definecolor{failbg}{RGB}{248, 230, 234}      % Light red-gray background
\definecolor{failframe}{RGB}{176, 36, 24}    % Steel red frame
\definecolor{failbadge}{RGB}{225, 151, 168}      % Dark red gray badge

\definecolor{successbg}{RGB}{239, 255, 229}   % - Mint Green
\definecolor{successframe}{RGB}{34, 139, 34}  % - Forest Green
\definecolor{successbadge}{RGB}{182, 200, 108}% - Olive/Light Green

\definecolor{ThemePurple}{RGB}{95, 50, 105}
\definecolor{ThemeBlue}{RGB}{32, 56, 100}
\definecolor{querycolor}{RGB}{112, 112, 112} 
\definecolor{verifycolor}{RGB}{117, 70, 126}
\definecolor{searchcolor}{RGB}{217, 140, 70}
\definecolor{solvercolor}{RGB}{217, 140, 70}

\hypersetup{
    colorlinks=true,
    linkcolor=ThemeBlue,   % ThemeBlue   MidnightBlue
    citecolor=ThemeBlue,   
    urlcolor=ThemeBlue,
    }

\usepackage{booktabs}
\usepackage{multirow}
\usepackage{makecell}
\usepackage{array}
\usepackage{siunitx}

\definecolor{TableBG}{RGB}{238, 240, 242}
\definecolor{DeltaUp}{RGB}{117, 70, 126}
\definecolor{DeltaDown}{RGB}{217, 140, 70}
\definecolor{DeltaZero}{RGB}{150, 150, 150}

\newtcolorbox{promptbox}[1][]{
  colback=gray!5!white,        % 浅灰背景
  colframe=black!75!black,     % 深黑边框
  boxrule=0.3mm,               % 边框宽度
  arc=3mm,                     % 圆角半径
  auto outer arc=true,         % 自动调整外圆角
  width=\linewidth,            % 宽度占满行
  fontupper=\footnotesize,            % 小号
  #1                           % 允许传入额外参数（如 title）
}

\tcbset{
    agentstyle/.style={
        colback=gray!5!white,          % 【保留】浅灰背景
        boxrule=0.3mm,                 % 【保留】边框宽度
        arc=3mm,                       % 【保留】圆角半径
        auto outer arc=true,           % 【保留】自动外圆角
        fontupper=\small               % 内容字体略小
    }
}

\newtcolorbox{verifierbox}[1][]{
    agentstyle,
    colframe=gray!60!verifycolor, % 使用 verifycolor
    title=\textbf{Verifier Agent},
    #1 % 允许传入额外参数
}

\newtcolorbox{searchbox}[1][]{
    agentstyle,
    colframe=gray!60!searchcolor, % 使用 searchcolor
    title=\textbf{Search Agent},
    #1
}

\newtcolorbox{answerbox}[1][]{
    agentstyle,
    colframe=teal!75!black,       % 这里用了您之前的 teal 颜色
    title=\textbf{Answer Agent},
    #1
}

\newtcolorbox{solverbox}[1][]{
    agentstyle,
    colframe=gray!60!solvercolor, 
    title=\textbf{Solver Agent},
    #1
}

\newtcolorbox{envbox}[1][]{
    agentstyle,
    colframe=gray!60!querycolor,
    title=#1, % 标题可变 (如 Query 或 Environment)
}

\usepackage{amsmath}
\usepackage{amssymb}
\usepackage{amsthm}
\usepackage[utf8]{inputenc}
\usepackage[T1]{fontenc}
\usepackage{url}
\usepackage{nicefrac}
\usepackage{microtype}
\usepackage{xcolor}
\usepackage{graphicx}
\usepackage{framed}
\usepackage{makecell}

\usepackage{caption}
\usepackage{subcaption}
\usepackage{wrapfig}
\usepackage{multirow}
\usepackage{amsfonts}
\usepackage{bm}
\usepackage{verbatim}
\usepackage{longtable}
\usepackage{arydshln}
\usepackage{colortbl}
\usepackage{hyperref}
\usepackage{bm} 
\usepackage{algorithm}
\usepackage{algpseudocode}
\usepackage{xspace}
\usepackage{booktabs}

\usepackage{array}      % m{...} 列格式、列对齐
\usepackage{adjustbox}  % 将表格宽度限制在页面范围内
\usepackage{colortbl}   % 表头、行和单元格背景色
\usepackage{multirow}   % 跨行单元格
\usepackage{makecell}   % 表头换行，以及 \Xhline 粗横线
\usepackage{bbding}     % 原 Table 3 的勾、叉符号
\usepackage{placeins}   % \FloatBarrier，限制表格浮动位置
\usepackage{pifont}
\usepackage{bm}
\usepackage{amsmath, amssymb} % 一般模板已有，确保有 \eqref 和 \emptyset
\usepackage{xcolor}           % 若 CVPR 模板已加载 xcolor，可省略这一行避免选项冲突

\definecolor{bestColor}{HTML}{faecf7} 
\definecolor{secondColor}{HTML}{e4f3fa}

\newcommand{\method}{{\fontfamily{lmtt}\selectfont EmboEagle}\xspace}
\title{EmboEagle: An Efficient Agentic Embodied Framework for Long-horizon Robotic Manipulation}

\author[1*]{Fan Xu}
\author[2*]{Zhongtao Wang}
\author[3]{Hao Wu}
\author[4]{Yitong Zhu}
\author[1]{Wenzhuo Ma}
\author[5]{Long Li}
\author[1]{Zhuohao Li}
\author[6]{Zhisheng Chen}
\author[1]{Rui Su}
\author[6]{Kun Wang}
\author[7\dag]{Xian Wu}
\author[1,8\dag]{Wanli Ouyang}

\renewcommand{\affiliationlist}{%
  \affiliationformat[1]{Shenzhen Loop Area Institute}, \affiliationformat[2]{Sichuan University},
  \affiliationformat[3]{Tsinghua University}, \\[2pt]
  \affiliationformat[4]{The Hong Kong University of Science and Technology (Guangzhou)}, \\[2pt]
  \affiliationformat[5]{Griffith University}, \affiliationformat[6]{Nanyang Technological University}, \\[2pt]
  \affiliationformat[7]{Tencent}, \affiliationformat[8]{The Chinese University of Hong Kong}%
}

\contribution[*]{Equal contribution $~~$}
\contribution[\dagger]{Corresponding Author}

\abstract{

Long-horizon robotic manipulation demands reliable execution across interdependent stages, yet standalone World-Action Models (WAMs) suffer from compounding physical errors, while stateless Vision-Language Model (VLM) supervision incurs prohibitive chunkwise latency and lacks 3D spatial persistence. We present \textbf{\method{}}, a closed-loop agentic framework that unifies subtask-conditioned WAM control, latent-space adaptive supervision, and memory-guided geometric recovery. First, we adapt sequential WAMs via decoupled two-stage flow matching that preserves cross-stage visual history while calibrating truncated action supervision at subtask boundaries. Second, without an auxiliary vision encoder, we distill teacher intervention decisions into a lightweight routing gate driven solely by the WAM Action Expert's pre- and post-chunk latent transitions and subtask embeddings. Third, upon invocation, the VLM decouples diagnostic status from discrete routing actions and executes parameterized recovery primitives grounded in structured memory (initial references, active grasp bindings, and tool traces), followed by mandatory post-repair verification. Across RMBench and LIBERO-Mem, \textbf{\method{}} achieves 91.3\% and 90.6\% average success, surpassing the strongest baselines by 8.0 and 9.8 percentage points, while boosting real-world success over VoLoAgent under identical executor weights. Moreover, adaptive gating cuts VLM calls by 74.2\% versus chunkwise monitoring (91.3\% vs. 93.7\% success).

}

\date{Nov 18, 2026}
\metadata[Author emails:]{\email{markxu@mail.ustc.edu.cn}}
\metadata[\faGithub~~Project:]{\url{https://github.com/Markus-xu/EmboEagle}}
\metadata[\ ]

\begin{document}
\maketitle

\newcommand{\fix}{\marginpar{FIX}}
\newcommand{\new}{\marginpar{NEW}}

\section{Introduction}
\label{sec:intro}

Long-horizon robotic manipulation requires a robot to preserve task intent across changing object configurations and grasp states, deciding when a subtask completes, retaining spatial cues needed by subsequent stages, and correcting physical errors before they compound. 
World-Action Models (WAMs) establish a strong foundation for this execution layer by coupling visual dynamics prediction with continuous action generation. Recent advances extend this paradigm through history-conditioned control in Long-WAM~\citep{huang2026longwam} and episodic visual guidance in MaP-WAM~\citep{map2026}. 
Nevertheless, turning a capable visuomotor executor into a reliable long-horizon agent raises a complementary control question: \emph{how should a system allocate high-level supervision and translate its decisions into physical corrections during an ongoing task?}

\begin{wrapfigure}[18]{r}{0.65\linewidth}
    \centering
    \vspace{-10pt}
    \includegraphics[width=1.0\linewidth]{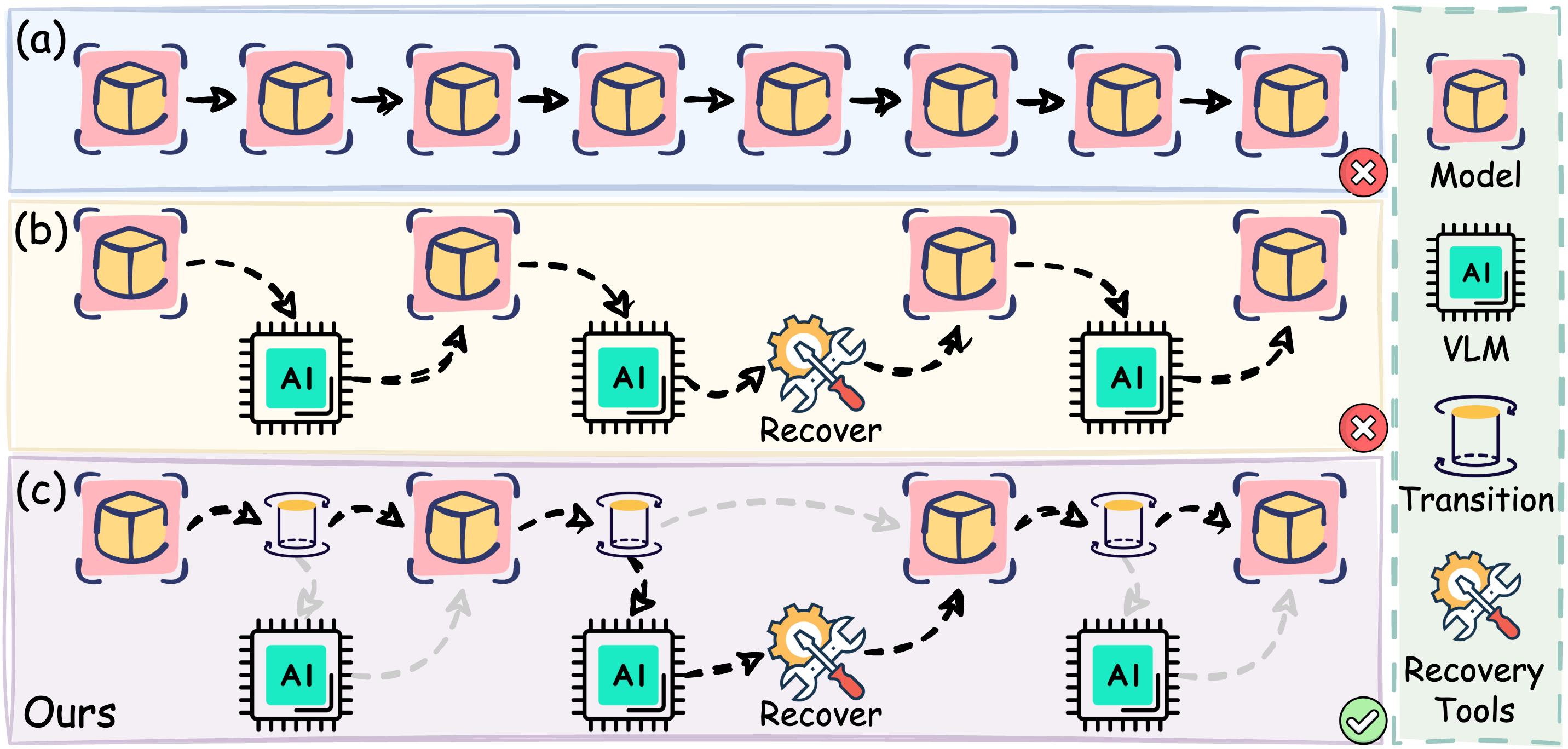}
    \caption{\textbf{Motivation of \method{}.} (a) Open-loop WAM execution suffers from unrecoverable errors. (b) Fixed-interval VLM supervision incurs redundant overhead on routine chunks. (c) \textbf{\method{}} uses a lightweight gate to trigger VLM orchestration and geometric recovery only when needed.}
    \label{fig:introduction}
    \vspace{-10pt}
\end{wrapfigure}

VLM-based orchestration bridges learned policies with explicit planning, monitoring, and manipulation tools~\citep{volo2026,gu2026harnesswam}, yet the timing of this supervision imposes a sharp efficiency--responsiveness trade-off. 
As contrasted in Figure~\ref{fig:introduction}(a--b), open-loop execution lacks error recovery, whereas querying a VLM after every action chunk repeatedly returns routine continuation commands and incurs unnecessary overhead; conversely, lengthening a fixed checking interval delays the detection of subtask completions, auxiliary tool opportunities, and failed grasps. 
While progress-based and event-driven triggers mitigate parts of this cost~\citep{gu2026harnesswam}, existing triggers often rely on heuristic rules or auxiliary visual processing. 
We instead investigate whether a lightweight predictor can use the executor's internal representations to anticipate a teacher's intervention decisions, including stage advancement, proactive assistance, and failure recovery.

Beyond knowing \emph{when} to intervene, effective recovery requires a persistent physical reference for \emph{what} to restore. 
Returning a displaced object to an earlier location requires retaining that location after the object has moved, and placing a held object requires accounting for the spatial offset between its geometric anchor and the gripper. 
Because the live observation, the initial scene reference, and the active grasp record each supply distinct geometric evidence, structured memory is essential both for semantic interpretation and for synthesizing corrective trajectories. 
Moreover, once a corrective motion finishes, fresh observations must be reassessed before low-level policy control safely resumes.

To address these challenges, we propose \textbf{\method{}}, an agentic framework that integrates subtask-conditioned WAM execution with adaptive VLM invocation and memory-guided geometric recovery (Figure~\ref{fig:introduction}(c)). 
Crucially, its design \emph{separates the decision to request supervision from the supervisory decision itself}, uniting three tightly coupled components:
% \ding{68}
\ding{95}~\textbf{\textit{Boundary-aware subtask WAM adaptation.}} Starting from pretrained Long-WAM checkpoints, we perform decoupled two-stage flow-matching adaptation (Video DiT LoRA followed by full Action DiT fine-tuning) on subtask-conditioned demonstrations, preserving cross-stage visual history while truncating action supervision at subtask boundaries via a calibrated fixed-horizon loss.
\ding{95}~\textbf{\textit{Latent-delta adaptive VLM invocation.}} Rather than duplicating visual encoding with an external perception network, we train a lightweight routing gate distilled from teacher decisions using the WAM Action Expert's pre- and post-chunk token transitions and subtask embeddings, keeping post-intervention verification outside optional gating.
\ding{95}~\textbf{\textit{Decoupled orchestration and memory-guided recovery.}}
When invoked, the VLM separates diagnostic status from discrete routing actions.
Assist operations can establish gripper--object bindings for subsequent recovery or directly support execution.
Recovery draws on these bindings, initial references, and recent tool feedback to select tool primitives, with immediate verification after each intervention.
Our main contributions are threefold:

% \begin{itemize}[leftmargin=1.3em]
\begin{itemize}[leftmargin=1em,itemsep=-0.1em]
\item[\ding{202}] \textbf{Closed-loop agentic architecture.} We present \textbf{\method{}}, which unifies subtask-conditioned WAM execution, decoupled VLM task orchestration, and persistent geometric memory with mandatory post-repair verification.

\item[\ding{203}] \textbf{Latent-space adaptive VLM invocation.} We formulate selective monitoring as teacher-decision distillation over WAM action-token transitions and subtask semantics, enabling lightweight runtime gating without an auxiliary vision backbone.

\item[\ding{204}] \textbf{Boundary-aware WAM adaptation and empirical validation.} We develop a two-stage subtask adaptation scheme to adapt to the long-horizon settings. We validate our method alongside closed-loop evaluations on RMBench and LIBERO-Mem.
\end{itemize}
\section{Related Work}

\textbf{World Action Models for Robotic Manipulation.}
World action models (WAMs) couple visual dynamics prediction with action generation to bring video pretraining into robot control.
LingBot-VA~\citep{li2026lingbotva} jointly learns frame prediction and policy execution through an autoregressive diffusion framework, while DreamZero~\citep{ye2026dreamzero} adapts a pretrained video diffusion backbone to joint video--action modeling.
Further, Fast-WAM~\citep{yuan2026fastwam} retains video co-training while omitting explicit future prediction at inference, reducing action-generation latency.
Long-WAM~\citep{huang2026longwam} studies how autoregressive video pretraining supports the use of extended observation histories and combines causal world-action adaptation with efficient execution.
MaP-WAM~\citep{map2026} organizes historical observations into episodic records and generates language and visual plans from memory. Then it conditions action and progress prediction on those plans.
These models provide predictive and temporal representations that support higher-level orchestration.

\textbf{Agentic Embodied Frameworks.}
Agentic embodied frameworks connect high-level reasoning to executable skills and use feedback to coordinate multi-stage tasks~\citep{liu2023reflect,huang2022inner}.
SayCan~\citep{ahn2022saycan} grounds language-based skill selection in learned affordance estimates.
VoLo~\citep{volo2026} extends physical orchestration to interruptible learned policies, perception models, and manipulation primitives, allowing the controller to switch tools when execution requires correction.
HarnessWAM~\citep{gu2026harnesswam} maintains scene beliefs and a task graph, constrains plans to the executor's capabilities, and combines outcome verification with local recovery.
State maintenance provides another axis of design: HiMe~\citep{hime2026} separates execution, working memory, and longer-term planning, whereas LAW~\citep{law2026} couples planning, status assessment, and action evaluation with hierarchical visual memory.

\textbf{VLM supervision.}
The cost of VLM supervision motivates separating frequent observation from expensive reasoning~\citep{gu2026safe}.
Checking every action chunk can repeat unchanged decisions, whereas fixed intervals cannot adapt to when transitions or failures occur.
Existing systems use different signals to schedule deliberation.
HarnessWAM~\citep{gu2026harnesswam} uses learned progress and completion cues to trigger its task manager.
Spotter~\citep{li2026spotter} employs a local Qwen screener to flag potential errors for a VLM judge, with bounded asynchronous supervision and checks following repair.
TUD~\citep{yang2026tud} triggers generalist reasoning from the cross-step dispersion of action re-predictions, using computations already performed by its dual-system architecture.

\section{Problem Definition}
\label{sec:problem_definition}

We consider language-conditioned robotic manipulation in a partially observable environment, where accomplishing a global instruction $G$ requires completing a sequence of interdependent subtasks.
At control step $t$, the robot receives multi-view RGB observations $o_t$ and proprioceptive state $q_t$, with calibrated depth available for geometric perception.
Let $h_t$ denote the interaction history of observations, robot states, and executed actions up to $t$; early observations often establish object identities or spatial relations required later in the task~\citep{map2026}.
We assume access to a pretrained World-Action Model and target-task demonstrations annotated with subtask instructions from a supported vocabulary $\mathcal{G}$.

Let $g_{k(t)} \in \mathcal{G}$ denote the active subtask instruction, whose index $k(t)$ may span multiple control chunks.
Conditioned on its retained observation history $h_t^{\mathrm{W}}$, proprioception $q_t$, and active instruction $g_{k(t)}$, the WAM policy $\pi_\theta$ generates an $H$-step action chunk:
\begin{equation}
    A_t = (a_t, \ldots, a_{t+H-1})
    \sim \pi_{\theta}\!\left(\cdot \mid h_t^{\mathrm{W}}, q_t, g_{k(t)}\right),
    \label{eq:wam_interface}
\end{equation}
where $a_t \in \mathbb{R}^{d}$ is a robot control action and $H$ is the chunk horizon.

Executing long-horizon tasks purely in open loop leads to compounding errors and unrecoverable failures, whereas querying a Vision-Language Model (VLM) at every chunk boundary incurs substantial latency on routine motion.
Our problem therefore requires coordinating low-level chunk execution with a selective VLM invocation decision $b_t \in \{0, 1\}$ and structured physical recovery: bypassing supervision ($b_t = 0$) during routine progress while invoking high-level orchestration ($b_t = 1$) at stage transitions or failures to reliably fulfill $G$ with minimal redundant VLM queries.
\section{Methodology}

\begin{figure*}[t]
    \centering
    \includegraphics[width=\textwidth]{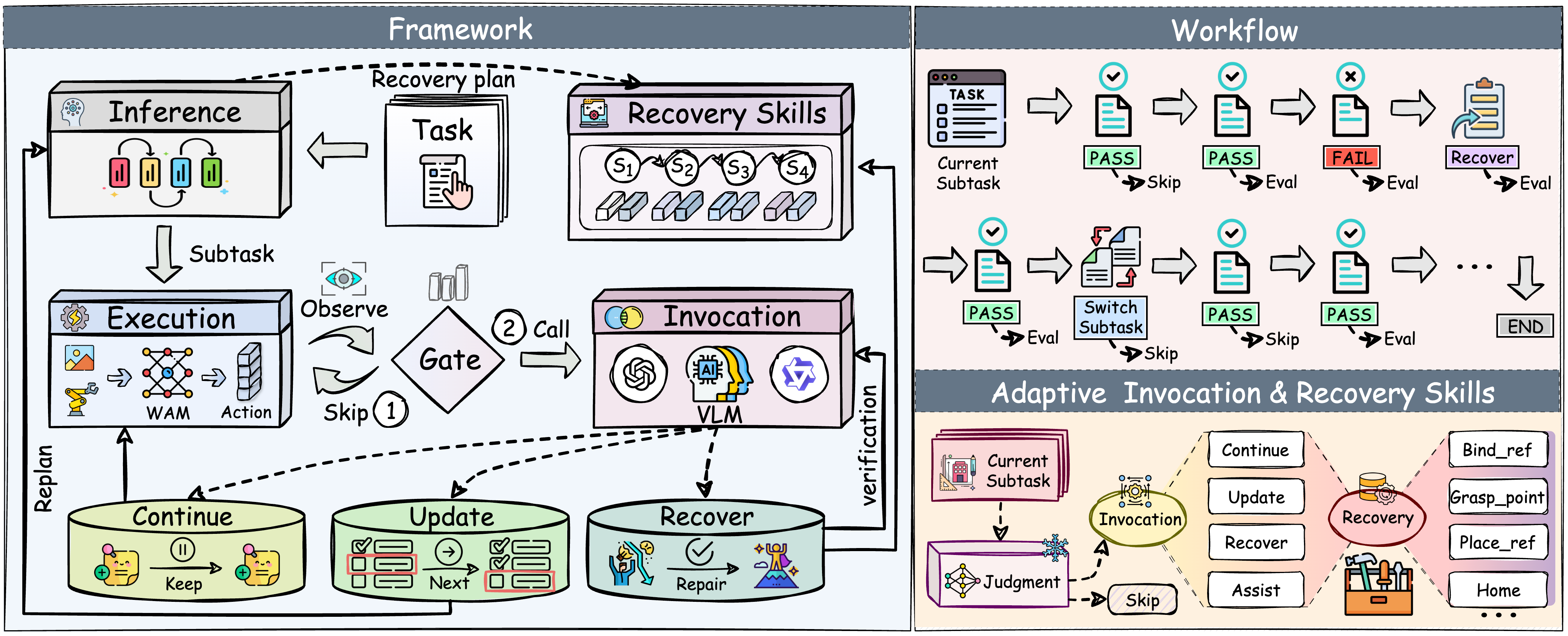}
    \vspace{-5pt}
    % \caption{Overview of \method{}. }
    \caption{\textbf{Overview of \method{}.} 
(\textit{Left}) Closed-loop architecture integrating WAM execution, gated VLM orchestration, and memory-guided recovery with post-repair verification. 
(\textit{Top Right}) Stepwise workflow bypassing routine chunks. 
(\textit{Bottom Right}) Latent-delta invocation gating and parameterized recovery skills.}
    \label{fig:framework}
    \vspace{-5pt}
\end{figure*}

\textbf{Overview.}
Figure~\ref{fig:framework} presents the overall closed-loop workflow of \textbf{\method{}}.
At runtime, a subtask-conditioned WAM executes chunkwise control while a structured memory anchors geometric invariants and a VLM orchestrates subtask transitions and physical recovery ($\triangleright$ Section~\ref{sec:framework}).
To eliminate redundant supervision during routine execution, a lightweight routing gate distilled from teacher decisions dynamically schedules VLM invocations from internal WAM transitions ($\triangleright$ Section~\ref{sec:adaptive_invocation}).
Finally, to support accurate stage execution and latent gating, we adapt the WAM through a boundary-aware two-stage flow-matching scheme ($\triangleright$ Section~\ref{sec:wam_adaptation}).

\subsection{The Framework of EmboEagle}
\label{sec:framework}

\textbf{High-level planning and structured state.}
\textbf{\method{}} organizes long-horizon execution through a hierarchical, tree-structured task plan rooted at the global goal $G$, decomposing the task into an ordered sequence of active and remaining subtasks $\mathcal{P}_t = (g_{k(t)}, \mathcal{P}_t^{\mathrm{rem}})$ that can be dynamically pruned or regenerated upon replanning.
Whenever high-level supervision is requested, the controller assembles a structured VLM state context:
\begin{equation}
    \mathcal{X}_t = \langle G,\; \mathcal{P}_t,\; \mathcal{V}_t,\; q_t,\; m_t \rangle,
    \label{eq:vlm_input}
\end{equation}
where $\mathcal{V}_t$ aggregates initial, current, and recent historical visual observations, $q_t$ provides end-effector poses and gripper openings, and $m_t$ denotes the structured recovery memory detailed below.

\textbf{Chunkwise execution.}
During routine control, the adapted WAM policy $\pi_\theta$ predicts action chunks $A_t$ via Eq.~\ref{eq:wam_interface} (Section~\ref{sec:wam_adaptation}), executed directly as joint or end-effector commands on simulated or real-world robotic arms.
To preserve physical consistency when resuming policy execution, active grasp bindings maintained in $m_t$ automatically enforce gripper closure, while single-arm instructions keep the non-active arm stationary.
Upon completing each chunk, the routing gate evaluates $b_t$ (Section~\ref{sec:adaptive_invocation}): on $b_t=0$, execution continues directly under $g_{k(t)}$; on $b_t=1$, the controller passes $\mathcal{X}_t$ to the VLM and pauses policy control.
Upon a subtask transition or tool intervention, the controller clears pending actions while retaining and continuously updating the visual history $h_t^{\mathrm{W}}$, allowing subsequent action predictions to incorporate the current subtask instruction and latest observations.

\textbf{VLM-based transition.}  
When queried, the VLM orchestrator $\mathcal{F}_{\mathrm{VLM}}$ maps $\mathcal{X}_t$ to a structured response $\mathcal{Y}_t = \langle s_t, u_t, \rho_t, \mathcal{R}_t \rangle$, cleanly decoupling the diagnostic status $s_t \in \{\texttt{ongoing},\allowbreak\ \texttt{achieved},\allowbreak\ \texttt{abnormal}\}$ from the executable decision $u_t \in \{\texttt{continue},\allowbreak\ \texttt{update},\allowbreak\ \texttt{recovery},\allowbreak\ \texttt{assist}\}$ (with textual justification $\rho_t$).
While $s_t$ supplies interpretable context for logging without overriding control routing, $u_t$ dispatches one of four distinct branches:
\begin{itemize}[leftmargin=1.2em, itemsep=2pt, topsep=3pt]
    \item[\ding{224}] \textbf{Continuation (\texttt{continue}):} Retains the active subtask instruction $g_{k(t)}$ for subsequent chunkwise execution.
    \item[\ding{224}] \textbf{Plan Update (\texttt{update}):} Either advances to the next subgoal or invokes an independent replanning query (via \texttt{update\_mode}) that synthesizes a validated replacement plan from $\mathcal{G}$ and resets the plan index, with a bounded end-of-plan check to catch remaining work.
    \item[\ding{224}] \textbf{Failure Recovery (\texttt{recovery}):} Emits a skill plan $\mathcal{R}_t$ to physically rectify detected execution failures.
    \item[\ding{224}] \textbf{Execution Assistance (\texttt{assist}):} Emits a skill plan $\mathcal{R}_t$ to proactively assist ongoing execution.
\end{itemize}

\textbf{Memory-guided geometric recovery.}
To ground \texttt{recovery} and \texttt{assist} interventions in 3D space, the controller maintains and encapsulates the structured memory $m_t$ into a compact JSON record:
\begin{equation}
    m_t = \big\{\,\mathtt{"refs"}:\mathcal{E}_0,\; \mathtt{"held"}:\mathcal{B}_t,\; \mathtt{"last\_tool"}:\mathcal{F}_t\,\big\},
    \label{eq:memory_record}
\end{equation}
where \textbf{\textit{(i) $\mathcal{E}_0$ (\texttt{refs})}} caches static 3D positions and planar orientations extracted from calibrated RGB-D observations at episode start; \textbf{\textit{(ii) $\mathcal{B}_t$ (\texttt{held})}} maintains per-arm object–gripper bindings, storing object anchor points in the gripper frame along with available relative orientation information; and \textbf{\textit{(iii) $\mathcal{F}_t$ (\texttt{last\_tool})}} logs the latest skill execution results and error traces.
Given a repair plan $\mathcal{R}_t = (\xi_1, \ldots, \xi_L)$, the controller sequentially executes parameterized primitives from a library of nine tools: motion and gripper resets (\texttt{lift}, \texttt{home}, \texttt{release}), memory grounding (\texttt{bind}, \texttt{bind\_reference}), and RGB-D manipulation (\texttt{grasp\_point}, \texttt{place\_ref}, \texttt{place\_point}, \texttt{press\_point}).
Each skill projects image targets into 3D world coordinates via calibrated depth or retrieves stored poses from \texttt{refs} and \texttt{held}, solving joint trajectories through inverse kinematics.
Crucially, both \texttt{recovery} and \texttt{assist} update $\mathcal{B}_t$ and $\mathcal{F}_t$ upon completion (or tool error), and immediately trigger a fresh observation-based VLM check before resuming WAM control.

\subsection{Adaptive VLM Invocation}
\label{sec:adaptive_invocation}

\textbf{Supervisory decision distillation.}
While monitoring every chunk reliably captures stage transitions and errors, most routine chunks simply continue the active subtask.
To avoid redundant queries, we roll out full chunkwise teacher supervision across task episodes to curate a dataset of $N$ chunk transitions paired with schema-valid VLM decisions, formulating invocation gating as binary classification (\textsc{Skip} vs.\ \textsc{Call}).
The binary supervision target $y_i \in \{0, 1\}$ is determined strictly by the teacher's top-level action $u_i$:
\begin{equation}
    y_i = \mathbb{I}\!\left[u_i \in \{\mathtt{update}, \mathtt{recovery}, \mathtt{assist}\}\right],
    \label{eq:teacher_labels}
\end{equation}
where $y_i = 0$ corresponds to $\mathtt{continue}$, completely independent of the diagnostic status $s_i$.

\textbf{Invocation prediction.}
Rather than introducing an auxiliary vision encoder to re-process post-chunk frames, our routing gate $f_\phi$ operates directly on internal latent transitions of the WAM's Action Expert conditioned on the active subtask.
For a chunk completed at step $t$, let $f_t^{-}, f_t^{+}$ denote the frozen Action Expert's final action-token features before and after chunk execution, and let $e_t$ be the cached subtask text embedding.
We compute the latent transition delta and predicted invocation logit via a lightweight MLP $f_\phi$:
\begin{equation}
    \Delta f_t = f_t^{+} - f_t^{-}, ~~
    z_t = \mathcal{N}(\Delta f_t, e_t), ~~
    r_t = f_\phi(z_t),
    \label{eq:gate_features}
\end{equation}
where $\mathcal{N}(\cdot)$ standardizes and normalizes each feature block using training-split statistics while keeping all WAM and text features frozen.
To handle task diversity, duplicate observations, and sparse \textsc{Call} events, we train $f_\phi$ with a weighted binary cross-entropy loss:
\begin{equation}
    \mathcal{L}_{\mathrm{gate}} =
    \frac{1}{N}\sum_{i=1}^{N} \omega_i
    \bigl[\operatorname{softplus}(r_i) - y_i r_i\bigr],
    \label{eq:gate_loss}
\end{equation}
where $\omega_i$ balances task and input frequencies while upweighting \textsc{Call} examples to prioritize intervention recall.
At deployment, the controller evaluates $b_t = \mathbb{I}[r_t \ge \eta]$ at regular chunk boundaries (where $\eta$ balances query savings against recall), while post-intervention verification and episode initialization bypass gating with mandatory VLM checks ($b_t = 1$).

\subsection{Subtask-Conditioned WAM Adaptation}
\label{sec:wam_adaptation}

\textbf{Subtask data curation.}
To enable stage-aligned chunk execution, we adapt a pretrained sequential WAM~\citep{huang2026longwam} to follow subtask instructions from the controller vocabulary $\mathcal{G}$ (optionally paired with global task or progress cues).
Demonstrations are partitioned by episode into contiguous subtask intervals $[t_s, t_e)$ prior to sampling training windows.
For any start step $t \in [t_s, t_e)$, the input visual history $h_t^{\mathrm{W}}$ spans across the subtask onset $t_s$ to preserve physical continuity across transitions (replicating the initial frame with a padding flag when history is insufficient at episode start), whereas future action supervision is truncated at $t_e$ via the valid-action mask $\mu_{t,j} = \mathbb{I}[t+j < t_e]$ for $j \in \{0, \ldots, H-1\}$.
Future visual supervision retains observations up to the completion state at $t_e$ (repeating and masking later frames), and boundary windows with fewer than $H$ valid actions are preserved to ensure responsive behavior near subtask endings.

\textbf{Two-stage adaptation.}
We adapt the backbone through two decoupled flow-matching stages~\citep{lipman2022flow} while keeping the proprioceptive, VAE, and text encoders frozen.
First, we equip the Video DiT's attention projections with LoRA adapters~\citep{hu2021lora} and optimize a future-latent flow-matching objective.
Second, we freeze the adapted video branch and fine-tune the full Action DiT under action flow matching: during training, the frozen video branch processes noised ground-truth future latents to supply visual key--value conditioning to the Action DiT, whereas at inference it synthesizes these future conditions directly from visual history.

To prevent naive per-window normalization ($1/N_t$) from disproportionately amplifying gradients on short truncated windows near subtask endings, we optimize a fixed-horizon calibrated action loss across noise levels $\sigma$:
\begin{equation}
        \mathcal{L}_{\mathrm{act}}
        = \mathbb{E}\!\left[
            w(\sigma)\,\frac{C}{H}
            \sum_{j=0}^{H-1}\mu_{t,j}\ell_{t,j}
        \right], 
        \quad
        \ell_{t,j} = \frac{1}{d}\big\|\hat{v}_{t,j} - v^*_{t,j}\big\|_2^2,
        \quad
        C = \frac{H}{\mathbb{E}_{t \sim p_{\mathrm{train}}}[N_t]},
    \label{eq:masked_action_loss}
\end{equation}
where $N_t = \sum_j \mu_{t,j}$, $\hat{v}_{t,j}$ and $v^*_{t,j}$ denote predicted and target action velocity flows, and $w(\sigma)$ is the noise-level weight.
Computed once from the training sampling distribution, $C$ preserves expected supervision scale across the dataset while eliminating short-window gradient distortion.

\section{Experiments}

\subsection{Experimental Settings}

\textbf{Datasets.}
We evaluate long-horizon, subtask-wise robotic manipulation on RMBench~\citep{chen2026rmbench} and LIBERO-Mem~\citep{chung2026rethinking} using full-task success rate.
RMBench comprises nine bimanual tasks, including five $M(1)$ tasks requiring a single task-relevant past observation and four $M(n)$ tasks requiring multiple observations.
LIBERO-Mem comprises ten single-arm tasks involving single placement, counting, and sequencing/spatial reasoning.
We further design two real-world tasks: \emph{garbage sorting}, which requires sorting colored objects into their corresponding containers, and \emph{bread toasting}, which involves processing two slices of bread through a toaster-and-plate sequence.
Both tasks require coordinating multiple manipulation stages while tracking which objects have already been processed.

\textbf{Baselines.}
We compare \method{} against four families of representative methods aligned with each evaluation suite.
\ding{68} \textbf{\textit{VLA-based policies}} include $\pi_{0.5}$~\citep{intelligence2025pi_}, X-VLA~\citep{zheng2026x}, MemoryVLA~\citep{shi2026memoryvla}, and ActMem-VLA~\citep{zhao2026remember}.
\ding{68} \textbf{\textit{WAM-based policies}} include HiMem-WAM~\citep{sun2026himem}, AR-WAM~\citep{jiang2026ar}, and MemoryVAM~\citep{jiang2026memoryvam}.
\ding{68} \textbf{\textit{Agentic frameworks}} include 2AM~\citep{hu20262am}, MaP-WAM~\citep{map2026}, and VoLoAgent~\citep{volo2026}.
\ding{68} \textbf{\textit{Other policy baselines}} include DP~\citep{chi2025diffusion}, SlotSSM~\citep{chung2026rethinking}, and Mem-0~\citep{chen2026rmbench} as visuomotor and memory-based policy references.
See Appendix A for detailed descriptions.

\textbf{Implementation details.}
We segment demonstration episodes into subtasks via the VLM planner and initialize executors from Long-WAM~\citep{huang2026longwam} RoboTwin2.0~\citep{chen2025robotwin} and LIBERO~\citep{liu2023libero} checkpoints (the same architecture for real-world tasks).
Policies condition on the active subtask for RMBench and additionally incorporate the global task and stage index for LIBERO-Mem, observing 48 steps of visual history to predict $H=32$ action chunks.
We train Video DiT LoRA adapters for 10k steps before fine-tuning the full Action DiT for 30k steps on four NVIDIA H100 80GB GPUs.
High-level orchestration uses \texttt{GPT-6-Astra}~\citep{zhang2026unexpected} as $\mathcal{F}_{\text{VLM}}$, while the MLP routing gate $f_\phi$ is trained on frozen transition features with split-wise standardization.

\begin{table*}[!t]
\centering
\caption{\textbf{Results on RMBench.} Each cell reports task-wise success rate (\%) across seven diverse baselines. \textbf{Bold} and \underline{underlined} entries indicate the best and second-best results in each row, respectively.}
\vspace{-5pt}
\label{tab:rmbench_results}
\small
\setlength{\tabcolsep}{4.0pt}
\renewcommand{\arraystretch}{1.2}
\begin{adjustbox}{max width=\textwidth}
\resizebox{1.0\textwidth}{!}{
% \begin{tabular}{l|cccccccc}
\begin{tabular}{l|ccccccc>{\columncolor{ThemeBlue!5}}c}
\Xhline{1.2pt}
% \rowcolor{CadetBlue!15}
\textbf{Tasks}
& \makecell[c]{\textbf{DP} }
& \makecell[c]{\boldmath$\pi_{0.5}$ }
& \makecell[c]{\textbf{X-VLA} }
& \makecell[c]{\textbf{Mem-0} }
& \makecell[c]{\textbf{HiMem-WAM} }
& \makecell[c]{\textbf{AR-WAM} }
& \makecell[c]{\textbf{MaP-WAM} }
& \makecell[c]{\textbf{EmboEagle}\textbf{(Ours)}} \\
\Xhline{1.2pt}
Observe and Pick Up & 1\% & 9\% & 9\% & 4\% & 28\%                     & \underline{84\%} & 19\%  & \textbf{89\%} \\
% \rowcolor{gray!10}
Rearrange Blocks & 0\% & 13\% & 13\% & \underline{89\%} & 33\%         & 86\% &   66\%   & \textbf{100\%} \\
Put Back Block & 0\% & 11\% & 18\% & 90\% & 32\%                       & 88\% & \textbf{100\%}   & \underline{95\%} \\
% \rowcolor{gray!10}
Swap Blocks & 11\% & 24\% & 16\% & 67\% & 38\%                         & 82\% & \textbf{97\%}  &  \underline{92\%} \\
Swap T & 20\% & 15\% & 3\% & 14\% & 27\%                               & 16\% &  \underline{96\%}   &  \textbf{100\%} \\
\hline
% \rowcolor{gray!10}
Battery Try & 10\% & 16\% & 26\% & 28\% & 28\%                         & 52\% & \textbf{82\%}   & \underline{66\%}  \\
Blocks Ranking Try & 10\% & 6\% & 1\% & 18\%  & 24\%                  & 34\% & \textbf{94\%}     &  \underline{92\%} \\
% \rowcolor{gray!10}
Cover Blocks & 0\% & 0\% & 2\% & 68\% &  19\%                          &  78\% & \textbf{100\%}     & \underline{93\%} \\
Press Button & 0\% & 0\% & 0\% & 0\% &  8\%                            &  90\%   & \textbf{96\%}    & \underline{95\%} \\
\hline
% \rowcolor{gray!10}
\textbf{Total Average} & 5.8\% & 10.4\% & 9.8\% & 42.0\% & 26.3\%      & 67.8\% & \underline{83.3\%}    & \textbf{91.3\%} \\
\Xhline{1.2pt}
\end{tabular}
}
\end{adjustbox}
% \vspace{-5pt}
\end{table*}

\begin{figure*}[t]
    \centering
    \includegraphics[width=1.0\linewidth]{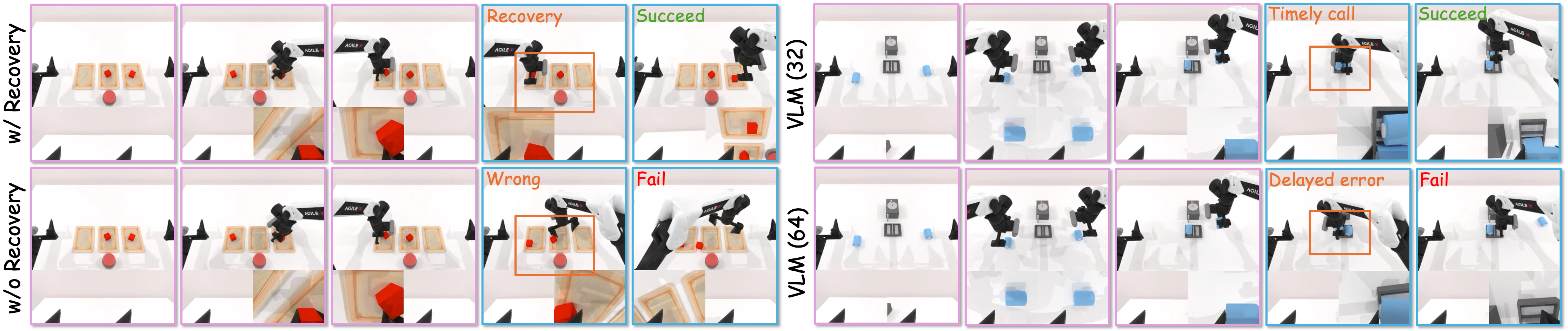}
    \vspace{-10pt}
    \caption{\textbf{Physical recovery and timely supervision.} \textbf{Left:} recovery rectifies an execution error, whereas disabling it leads to failure. \textbf{Right:} 32-step checks catch errors in time, while a 64-step interval delays intervention. Orange boxes mark key scene changes.}
    \label{fig:exp1}
    \vspace{-5pt}
\end{figure*}

\subsection{Main Results}

\textbf{RMBench.}
Across the nine memory-dependent bimanual tasks in Table~\ref{tab:rmbench_results}, \method{} achieves 91.3\% mean success, outperforming the strongest baseline MaP-WAM (83.3\%) by 8.0 points and AR-WAM (67.8\%) by 23.5 points.
Crucially, \method{} trains a single unified model across all tasks, whereas MaP-WAM trains a separate model per task.
While flat visuomotor and VLA policies (DP, $\pi_{0.5}$, X-VLA) stay below 11\%, and memory-augmented baselines suffer sharp task-specific drops (e.g., MaP-WAM 19\% on Observe and Pick Up; HiMem-WAM 8\% on Press Button; AR-WAM 16\% on Swap T), \method{} ranks top-two on all nine tasks (first on three) and reaches at least 89\% on eight.
Across memory regimes, \method{} attains 95.2\% on $M(1)$ tasks, substantially ahead of MaP-WAM (75.6\%) and AR-WAM (71.2\%), while maintaining 86.5\% on $M(n)$ tasks (versus 63.5\% for AR-WAM and 19.8\% for HiMem-WAM).
On early-observation and stage-handoff tasks such as Observe and Pick Up (89\% vs.\ 19\% for MaP-WAM) and Rearrange Blocks (100\% vs.\ 66\%), closed-loop orchestration and 3D-memory grounding prevent early drift from compounding (Figure~\ref{fig:exp1}, left).
Even on contact-sensitive Battery Try, \method{} scores 66\% (second to MaP-WAM at 82\%, and 14 points above AR-WAM), confirming consistent cross-task reliability.

\textbf{LIBERO-Mem.}
On the ten single-arm tasks of LIBERO-Mem (Table~\ref{tab:libero_mem_results}), \method{} achieves 90.6\% average success, improving over the strongest baseline ActMem-VLA (80.8\%) by 9.8 percentage points and the agentic framework 2AM (76.3\%) by 14.3 points, while ranking first or tied first on seven tasks.
On single-placement (\texttt{P}) and counting (\texttt{C}) tasks, \method{} achieves 100\% and 96.8\% average success, respectively.
Notably, implicit-memory policies degrade sharply as repetition counts grow: on Bowl lift-and-return $\times 5$ and $\times 7$, MemoryVLA drops to 3\% and 0\%, and $\pi_{0.5}$ falls from 82\% to 35\%, whereas \method{} maintains 100\% and 98\% by explicitly tracking progress in the hierarchical task plan $\mathcal{P}_t$.
On sequencing and spatial tasks (\texttt{S}), \method{} preserves high precision on basket centering (98\% and 96\%, versus 60\% and 48\% for 2AM) and establishes clear leads on multi-object permutations, outperforming all baselines by 16 points on Swap two bowls (58\% vs.\ 42\%) and 12 points on Rotate three bowls (67\% vs.\ 55\%), though visually symmetric object permutations remain the primary residual bottleneck.

\begin{table*}[!t]
\centering
\caption{\textbf{Results on Libero-Mem.} Each cell reports task-wise success rate (\%). Columns cover single placement (P), counting (C), and sequencing/spatial tasks (S). \textbf{Bold} and \underline{underlined} entries indicate the best and second-best results in each row, respectively.}
% \vspace{-5pt}
\label{tab:libero_mem_results}
\small
\setlength{\tabcolsep}{4.0pt}
\renewcommand{\arraystretch}{1.2}
\begin{adjustbox}{max width=\textwidth}
\resizebox{1.0\textwidth}{!}{
% \begin{tabular}{l|cccccccc}
\begin{tabular}{l|ccccccc>{\columncolor{ThemeBlue!5}}c}
\Xhline{1.2pt}
% \rowcolor{CadetBlue!15}
\textbf{Tasks} & Type
& \makecell[c]{\textbf{SlotSSM} }
& \makecell[c]{\boldmath$\pi_{0.5}$ }
& \makecell[c]{\textbf{MemoryVLA} }
& \makecell[c]{\textbf{2AM} }
& \makecell[c]{\textbf{MemoryVAM} }
& \makecell[c]{\textbf{ActMem-VLA} }
& \makecell[c]{\textbf{EmboEagle}\textbf{(Ours)}} \\
\Xhline{1.2pt}
Bowl lift-and-return & P                   & 50\% & 77\% & \underline{98\%} & \underline{98\%} & 75\% & 93\% & \textbf{100\%} \\
Bottle lift-and-return & P                 & 0\% & 90\% & 88\% & \textbf{100\%} & 65\% & \underline{93\%} & \textbf{100\%} \\
\hline
Bowl lift-and-return $\times 3$ & C        & 33\% & 52\% & 15\% & \textbf{96\%} & 65\% & 81\% & \underline{92\%} \\
Bottle lift-and-return $\times 3$ & C      & 0\% & 40\% & 20\% & \underline{86\%} & 15\% & \underline{86\%} & \textbf{97\%} \\
Bowl lift-and-return $\times 5$ & C        & 14\% & 82\% & 3\% & \underline{97\%} & 55\% & 95\% & \textbf{100\%} \\
Bowl lift-and-return $\times 7$ & C        & 0\% & 35\% & 0\% & \underline{96\%} & 35\% & 76\% & \textbf{98\%} \\
\hline
Swap two bowls & S                         & 0\% & 25\% & 33\% & \underline{42\%} & 10\% & 33\% & \textbf{58\%} \\
Rotate three bowls & S                     & 0\% & 52\% & 53\% & 41\% & 10\% & \underline{55\%} & \textbf{67\%} \\
Center occupied basket & S                 & 30\% & \textbf{100\%} & 92\% & 60\% & 25\% & 97\% & \underline{98\%} \\
Center empty basket & S                    & 20\% & \textbf{100\%} & 92\% & 48\% & 50\% & \underline{99\%} & 96\% \\
\hline
\textbf{Total Average} & -                 & 14.8\% & 65.2\% & 49.5\% & 76.3\% & 40.5\% & \underline{80.8\%} & \textbf{90.6\%} \\
\Xhline{1.2pt} 
\end{tabular}
}
\end{adjustbox}
% \vspace{-5pt}
\end{table*}

\begin{figure*}[t]
    \centering
    \includegraphics[width=1.0\linewidth]{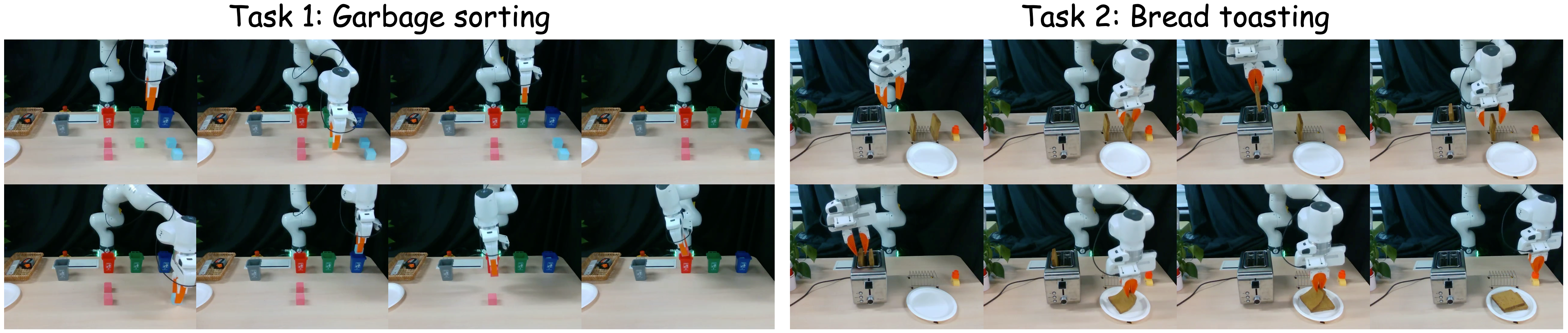}
    % \vspace{-5pt}
    \caption{\textbf{Real-world task demonstrations.} Example sequences for garbage sorting (left) and bread toasting (right). The tasks involve repeated pick-and-place stages with task-dependent object and destination selection. These images show demonstration trajectories.}
    \label{fig:exp2}
    % % \vspace{-5pt}
\end{figure*}

\begin{wraptable}{r}{0.4\textwidth}
% \vspace{-10pt}
\renewcommand{\arraystretch}{1.3}
  \centering
  \caption{\textbf{Real-world framework comparison.} Success rates (\%) using the same WAM executor weights for both frameworks.}
  \vspace{-5pt}
  \label{tab:real_world_results}
  % \begin{sc}
  \resizebox{\linewidth}{!}{%
    \begin{tabular}{lcc}
    \toprule
    \textbf{Framework} & \makecell{\textbf{Garbage} \textbf{sorting}} & \makecell{\textbf{Bread} \textbf{toasting}} \\
    \midrule
    VoLoAgent & 32\% & 51\% \\
    \rowcolor{ThemeBlue!5}
    \textbf{Ours} & \textbf{45\%} & \textbf{67\%} \\
    \bottomrule
    \end{tabular}
  }
  % \end{sc}
\vspace{-10pt}
\end{wraptable}

\textbf{Real-World Evaluation.}
To isolate orchestration and recovery gains from low-level policy capacity, we compare \method{} against VoLoAgent~\citep{volo2026} using identical WAM executor weights on both physical tasks (Figure~\ref{fig:exp2}).
As shown in Table~\ref{tab:real_world_results}, \method{} substantially outperforms VoLoAgent on both garbage sorting and bread toasting.
By anchoring targets in persistent 3D memory and enforcing post-repair visual verification before resuming WAM control, \method{} reliably recovers from real-world sensing and actuation errors across stage transitions.

\subsection{Efficiency Analysis}

\begin{figure*}[!t]
    \centering
    \includegraphics[width=1.0\linewidth]{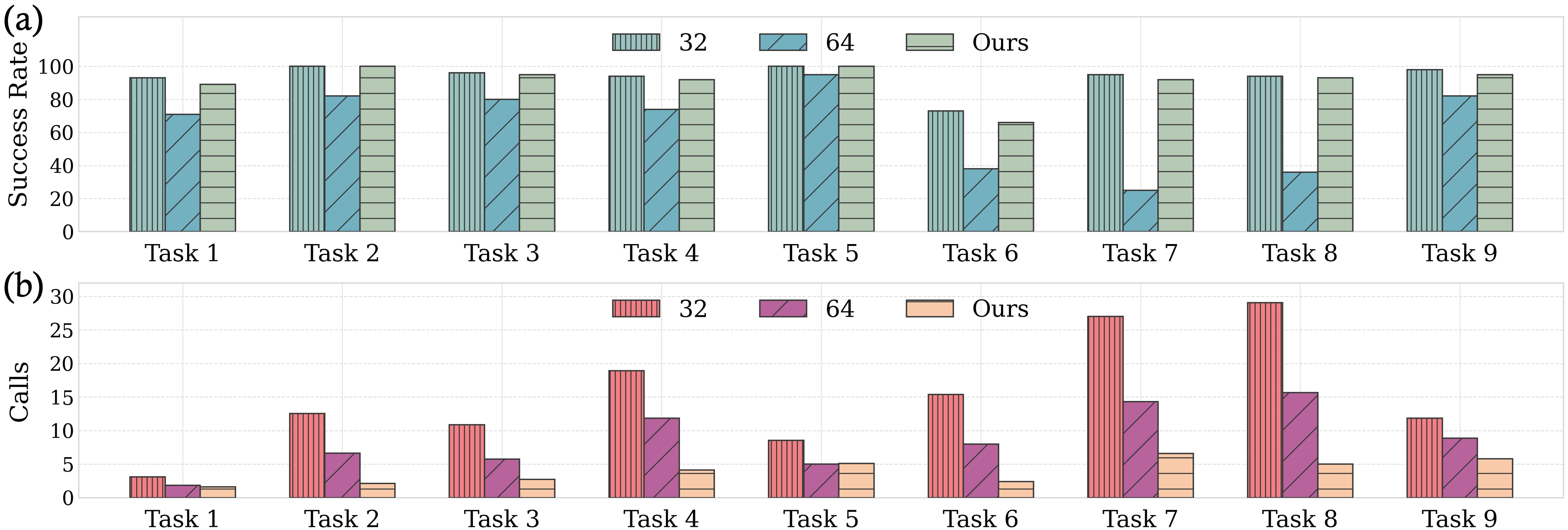}
    % \vspace{-5pt}
    \caption{\textbf{Invocation scheduling on RMBench.} (a) Full-task success rate for VLM monitoring every 32 or 64 control steps and adaptive invocation (Ours). (b) Mean monitoring-call counts per successful episode. Tasks 1–9 follow the row order of Table~\ref{tab:rmbench_results}.}
    \label{fig:exp3}
    % \vspace{-5pt}
\end{figure*}

To evaluate the accuracy-efficiency trade-off of adaptive VLM invocation, we compare our learned routing gate against fixed 32-step (chunkwise) and 64-step monitoring schedules across the nine RMBench tasks.
As shown in Figure~\ref{fig:exp3}(a--b), adaptive invocation achieves 91.3\% mean success with only 3.93 VLM calls per successful episode, remaining within 2.4 percentage points of dense 32-step supervision (93.7\%) while cutting query demand by 74.2\% (from 15.26 calls).
Moreover, adaptive scheduling strictly dominates the 64-step fixed baseline in both metrics, improving mean success by 26.5 percentage points (91.3\% vs.\ 64.8\%) with 54.6\% fewer queries (3.93 vs.\ 8.66).
Uniformly lengthening the interval misses consequential transitions and delays repair, depressing 64-step success on Blocks Ranking Try (25\% vs.\ 92\%) and Cover Blocks (36\% vs.\ 93\%), consistent with Figure~\ref{fig:exp1} (right).
By contrast, our gate allocates supervision dynamically to match task demand, dropping to 2.1 calls on routine Rearrange Blocks while preserving 5.1 calls on Swap T.

\subsection{Core Design Ablation}

\begin{wraptable}{r}{0.5\textwidth}
\vspace{-10pt}
\renewcommand{\arraystretch}{1.3}
  \centering
  \caption{Performance of ablation study on RMBench. \textbf{Bold} and \underline{underlined} entries indicate the best and second-best.}
  \vspace{-5pt}
  \label{tab:ablation}
  % \begin{sc}
  \resizebox{\linewidth}{!}{%
    \begin{tabular}{l cc cc}
    \toprule
    \textbf{Variants} & \textbf{swap\_T} & \textbf{put\_back\_b} & \textbf{rearrange\_b} & \textbf{cover\_b} \\
    \midrule
    WAM only & 5\% & 11\% & 8\% & 0\% \\
    w/o Recovery & 49\% & 58\% & 64\% & 47\% \\
    w/o Assist & \underline{84\%} & \underline{91\%} & \underline{98\%} & \underline{90\%} \\
    \rowcolor{ThemeBlue!5}
    \textbf{Ours} & \textbf{100\%} & \textbf{95\%} & \textbf{100\%} & \textbf{93\%} \\
    \bottomrule
    \end{tabular}
  }
  % \end{sc}
\vspace{-10pt}
\end{wraptable}

To isolate each orchestration mechanism, we evaluate three ablated variants of \method{} on four RMBench tasks using full-task success rate:
\ding{182}~\textbf{WAM only}, which runs the adapted WAM without VLM orchestration, 3D memory, or skill interventions;
\ding{183}~\textbf{w/o Recovery}, which disables reactive \texttt{recovery} (and post-repair verification) while keeping stage updates and \texttt{assist}; and
\ding{184}~\textbf{w/o Assist}, which removes proactive \texttt{assist} during ongoing execution while preserving stage updates and \texttt{recovery}.

Table~\ref{tab:ablation} shows that removing any module degrades task completion.
\emph{WAM only} collapses to 6.0\% mean success (0\% on Cover Blocks), confirming that local chunk execution alone fails over multi-stage horizons.
Disabling \texttt{recovery} (\emph{w/o Recovery}) lowers mean success by 42.5 points (97.0\% to 54.5\%), with the steepest drops on Swap T (100\% to 49\%) and Cover Blocks (93\% to 47\%), where unrectified slips or misalignments cascade into failure (Figure~\ref{fig:exp1}, left).
Removing \texttt{assist} (\emph{w/o Assist}) reduces mean success by 6.2 points to 90.8\%, affecting Swap T (100\% to 84\%) far more than Put Back Block (95\% to 91\%) or Rearrange Blocks (100\% to 98\%).
Overall, \texttt{assist} curbs nascent drift in precision stages, while \texttt{recovery} ensures a 3D-memory-anchored fallback.

\subsection{Visualization and Analysis}
\label{sec:routing_analysis}

\begin{figure}[t]
    \centering
    \includegraphics[width=1.0\linewidth]{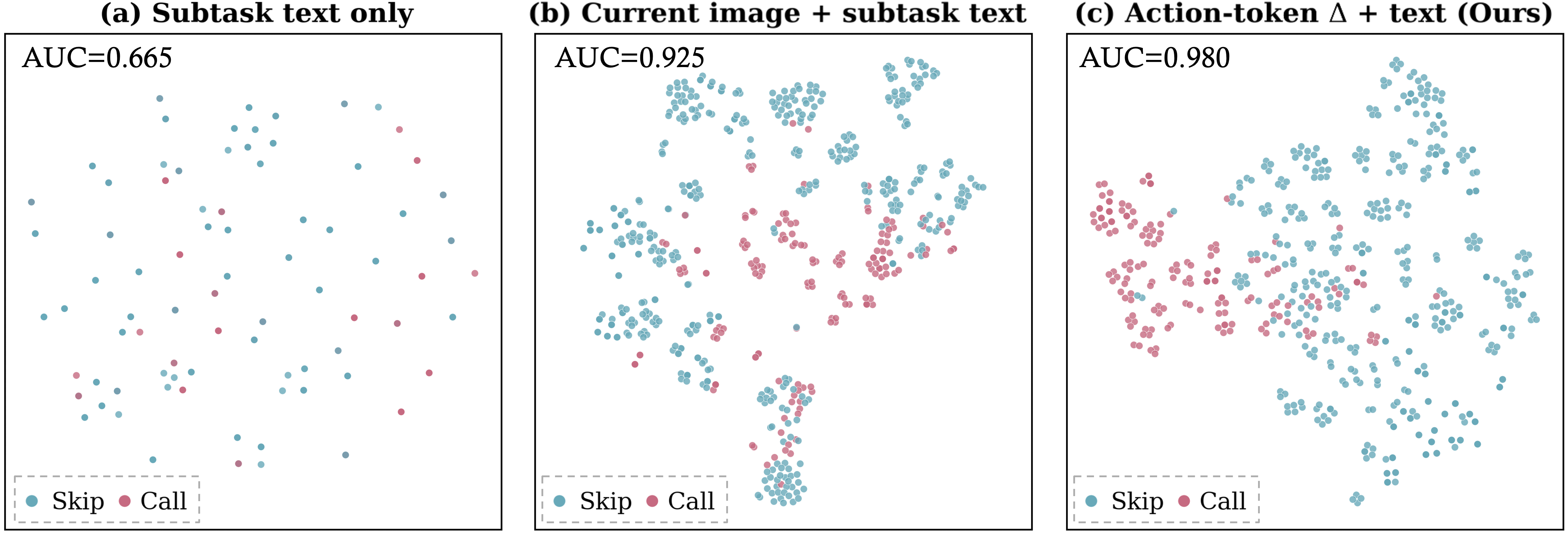}
    \vspace{-10pt}
    \caption{t-SNE of gate features using (a) text, (b) image + text, and (c) latent $\Delta f_t$ + text (\textsc{Skip}: blue; \textsc{Call}: pink).}
    \label{fig:exp4}
    \vspace{-5pt}
\end{figure}

To examine how internal policy representations support selective VLM invocation, we analyze the latent transitions of the frozen Action Expert.
At each chunk boundary, the pre-execution state $f_t^{-}$ provides a reference baseline for interpreting the post-chunk representation $f_t^{+}$.
Computing the action-token difference $\Delta f_t = f_t^{+} - f_t^{-}$ (Eq.~\ref{eq:gate_features}) suppresses static scene features that remain invariant across the chunk while preserving directional dynamics relative to the active subtask $e_t$, distinguishing routine progress from stalls or stage transitions.

Figure~\ref{fig:exp4} compares t-SNE~\citep{maaten2008visualizing} projections and gate AUC across three input configurations.
Using subtask text alone (Figure~\ref{fig:exp4}(a)) mixes \textsc{Skip} and \textsc{Call} thoroughly (0.665 AUC), as instructions provide no progress feedback.
Adding current-frame image features (Figure~\ref{fig:exp4}(b)) forms task-level clusters and raises AUC to 0.925, yet \textsc{Skip} and \textsc{Call} remain entangled within clusters because static frames miss execution dynamics.
In contrast, combining $\Delta f_t$ with subtask text (Figure~\ref{fig:exp4}(c)) concentrates \textsc{Call} instances into a more separable region despite modest local overlap, pushing AUC to 0.980 and validating latent transitions as an effective, encoder-free routing signal.

\section{Conclusion}

We presented \method{}, a closed-loop agentic framework for long-horizon robotic manipulation.
Using decoupled two-stage flow matching, \method{} retains cross-stage visual history while calibrating boundary-truncated supervision for subtask-aligned control.
A lightweight gate distilled from teacher decisions triggers VLM calls solely from the Action Expert's latent transitions and subtask embeddings without auxiliary encoders.
When invoked, the VLM decouples diagnostic status from routing actions and grounds recovery primitives in persistent geometric memory with post-repair verification.
Jointly, these mechanisms unify execution-driven supervision timing with anchored physical repair across multi-stage tasks.

% \clearpage
{\small
\bibliography{reference}

@article{li2026lingbotva,
  author = {Li, Lin and Zhang, Qihang and Luo, Yiming and Yang, Shuai and Wang, Ruilin and Han, Fei and Yu, Mingrui and Gao, Zelin and Xue, Nan and Zhu, Xing and Shen, Yujun and Xu, Yinghao},
  title = {{Causal World Modeling for Robot Control}},
  journal = {arXiv preprint arXiv:2601.21998},
  year = {2026},
  eprint = {2601.21998},
  archivePrefix = {arXiv},
  url = {https://arxiv.org/abs/2601.21998v1}
}

@article{ye2026dreamzero,
  author = {Ye, Seonghyeon and Ge, Yunhao and Zheng, Kaiyuan and Gao, Shenyuan and Yu, Sihyun and Kurian, George and Indupuru, Suneel and Tan, You Liang and Zhu, Chuning and Xiang, Jiannan and Malik, Ayaan and Lee, Kyungmin and Liang, William and Ranawaka, Nadun and Gu, Jiasheng and Xu, Yinzhen and Wang, Guanzhi and Hu, Fengyuan and Narayan, Avnish and Bjorck, Johan and Wang, Jing and Kim, Gwanghyun and Niu, Dantong and Zheng, Ruijie and Xie, Yuqi and Wu, Jimmy and Wang, Qi and Julian, Ryan and Xu, Danfei and Du, Yilun and Chebotar, Yevgen and Reed, Scott and Kautz, Jan and Zhu, Yuke and Fan, Linxi "Jim" and Jang, Joel},
  title = {{World Action Models are Zero-shot Policies}},
  journal = {arXiv preprint arXiv:2602.15922},
  year = {2026},
  eprint = {2602.15922},
  archivePrefix = {arXiv},
  url = {https://arxiv.org/abs/2602.15922v1}
}

@article{huang2026longwam,
  author = {Huang, Wei and Zhang, Bohan and Liu, Chenzhi and Liu, Isabella and Yang, Shuai and Mao, Weian and Wang, Luozhou and Xiao, Yicheng and Lin, Weifeng and Hu, Qixin and Chu, Bryan and Liu, Sifei and Fan, Linxi and Qi, Xiaojuan and Han, Song and Chen, Yukang},
  title = {{Long-WAM: Scaling the Context of World-Action Models}},
  journal = {arXiv preprint arXiv:2610.10528},
  year = {2026},
  eprint = {2610.10528},
  archivePrefix = {arXiv},
  url = {https://arxiv.org/abs/2610.10528v1}
}

@article{map2026,
  author = {Zhao, Sizhe and Xie, Haozhe and Zhao, Weiyu and Zhang, Chenchu and Wang, Huan and Wang, Chenyang and Liu, Qinglin and Zhang, Shengping},
  title = {{Memory as Plans: World-Action Modeling with Memory-Grounded Planning}},
  journal = {arXiv preprint arXiv:2609.11561},
  year = {2026},
  eprint = {2609.11561},
  archivePrefix = {arXiv},
  url = {https://arxiv.org/abs/2609.11561v1}
}

@article{ahn2022saycan,
  author = {Ahn, Michael and Brohan, Anthony and Brown, Noah and Chebotar, Yevgen and Cortes, Omar and David, Byron and Finn, Chelsea and Gopalakrishnan, Keerthana and Hausman, Karol and Herzog, Alex and Ho, Daniel and Hsu, Jasmine and Ibarz, Julian and Ichter, Brian and Irpan, Alex and Jang, Eric and Ruano, Rosario Jauregui and Jeffrey, Kyle and Jesmonth, Sally and Joshi, Nikhil J and Julian, Ryan and Kalashnikov, Dmitry and Kuang, Yuheng and Lee, Kuang-Huei and Levine, Sergey and Lu, Yao and Luu, Linda and Parada, Carolina and Pastor, Peter and Quiambao, Jornell and Rao, Kanishka and Rettinghouse, Jarek and Reyes, Diego and Sermanet, Pierre and Sievers, Nicolas and Tan, Clayton and Toshev, Alexander and Vanhoucke, Vincent and Xia, Fei and Xiao, Ted and Xu, Peng and Xu, Sichun and Yan, Mengyuan},
  title = {{Do As I Can, Not As I Say: Grounding Language in Robotic Affordances}},
  journal = {arXiv preprint arXiv:2204.01691},
  year = {2022},
  eprint = {2204.01691},
  archivePrefix = {arXiv},
  url = {https://arxiv.org/abs/2204.01691v1}
}

@article{volo2026,
  author = {Chen, Siyi and Hadfield, Hugo and Zook, Alex and Uy, Mikaela Angelina and Song, Chan Hee and Coumans, Erwin and Yang, Xuning and Ladhak, Faisal and Qu, Qing and Birchfield, Stan and Tremblay, Jonathan and Blukis, Valts},
  title = {{VoLo: A Physical Orchestrator for Open-Vocabulary Long-Horizon Manipulation}},
  journal = {arXiv preprint arXiv:2606.07723},
  year = {2026},
  eprint = {2606.07723},
  archivePrefix = {arXiv},
  url = {https://arxiv.org/abs/2606.07723v1}
}

@article{gu2026harnesswam,
  author = {Gu, Zhaopeng and Zhu, Bingke and Lin, Tianxi and Zhu, Guibo and Chen, Yingying and Wang, Kai and Yuan, Tingyu and Yang, Fan and Zhao, Chaoyang and Li, Zhaowen and Su, Peng and Wang, Jinqiao},
  title = {{HarnessWAM: Bridging Prediction and Deliberation in World Action Models}},
  journal = {arXiv preprint arXiv:2608.09516},
  year = {2026},
  eprint = {2608.09516},
  archivePrefix = {arXiv},
  url = {https://arxiv.org/abs/2608.09516v2}
}

@article{hime2026,
  author = {Ji, Li and Wang, Siyin and Qian, Pengfang and Yu, Xiaopeng and Tian, Yihai and Fei, Zhaoye and Gong, Jingjing and Qiu, Xipeng},
  title = {{HiMe: Hierarchical Embodied Memory for Long-Horizon Vision-Language-Action Control}},
  journal = {arXiv preprint arXiv:2607.03449},
  year = {2026},
  eprint = {2607.03449},
  archivePrefix = {arXiv},
  url = {https://arxiv.org/abs/2607.03449v1}
}

@article{li2026spotter,
  author = {Li, Long and Zhao, Qichao and Yang, Yue and Xu, Fan and Wang, Zhe and Liew, Alan Wee-Chung and Qu, Chao and Shen, Heng Tao and Pan, Shirui},
  title = {{Spotter: Let the Embodied Model Lead, and the VLM Reflect for It}},
  journal = {arXiv preprint arXiv:2609.36808},
  year = {2026},
  eprint = {2609.36808},
  archivePrefix = {arXiv},
  url = {https://arxiv.org/abs/2609.36808v1}
}

@article{yang2026tud,
  author = {Yang, Hyemin and Jeong, Wooseong and Lee, Giwon and Yoon, Kuk-Jin},
  title = {{Triggering Generalist Reasoning via Predictive Uncertainty for Dual-System VLA}},
  journal = {arXiv preprint arXiv:2610.05025},
  year = {2026},
  eprint = {2610.05025},
  archivePrefix = {arXiv},
  url = {https://arxiv.org/abs/2610.05025v2}
}

@article{law2026,
  author = {Cao, Jiajun and Tang, Kai and Zhang, Bohua and Qian, Zezhong and Li, Ying and Ju, Xiaozhu and Wei, Xiaobao and Liu, Jiaming and Ding, Yang and Chen, Sixiang and Sun, Yiding and Qian, Siyuan and Wang, Tinghao and Wu, Jixian and Guo, Yichen and Zhang, Yi and Zou, Shilong and Dai, Yong and Tang, Jian and Zhang, Shanghang},
  title = {{LAW: Unifying Language, Action, and World Modeling for Agentic Physical Intelligence}},
  journal = {arXiv preprint arXiv:2610.23334},
  year = {2026},
  eprint = {2610.23334},
  archivePrefix = {arXiv},
  url = {https://arxiv.org/abs/2610.23334v1}
}

@article{hu2021lora,
  author = {Hu, Edward J. and Shen, Yelong and Wallis, Phillip and Allen-Zhu, Zeyuan and Li, Yuanzhi and Wang, Shean and Chen, Weizhu},
  title = {{LoRA: Low-Rank Adaptation of Large Language Models}},
  journal = {arXiv preprint arXiv:2106.09685},
  year = {2021},
  eprint = {2106.09685},
  archivePrefix = {arXiv},
  url = {https://arxiv.org/abs/2106.09685v1}
}

@article{yuan2026fastwam,
  author = {Yuan, Tianyuan and Dong, Zibin and Liu, Yicheng and Zhao, Hang},
  title = {{Fast-WAM: Do World Action Models Need Test-time Future Imagination?}},
  journal = {arXiv preprint arXiv:2603.16666},
  year = {2026},
  eprint = {2603.16666},
  archivePrefix = {arXiv},
  url = {https://arxiv.org/abs/2603.16666v1}
}

@article{chi2025diffusion,
  title={Diffusion policy: Visuomotor policy learning via action diffusion},
  author={Chi, Cheng and Xu, Zhenjia and Feng, Siyuan and Cousineau, Eric and Du, Yilun and Burchfiel, Benjamin and Tedrake, Russ and Song, Shuran},
  journal={The International Journal of Robotics Research},
  volume={44},
  number={10-11},
  pages={1684--1704},
  year={2025},
  publisher={Sage Publications, Sage UK: London, England}
}

@article{intelligence2025pi_,
  title   = {{${\pi}_{0.5}$}: A Vision-Language-Action Model with Open-World Generalization},
  author  = {Physical Intelligence and Black, Kevin and Brown, Noah and Darpinian, James and Dhabalia, Karan and Driess, Danny and Esmail, Adnan and Equi, Michael and Finn, Chelsea and Fusai, Niccolo and others},
  journal = {arXiv preprint arXiv:2504.16054},
  year    = {2025}
}

@inproceedings{zheng2026x,
  title={X-vla: Soft-prompted transformer as scalable cross-embodiment vision-language-action model},
  author={Zheng, Jinliang and Li, Jianxiong and Wang, Zhihao and Liu, Dongxiu and Kang, Xirui and Feng, Yuchun and Zheng, Yinan and Zou, Jiayin and Chen, Yilun and Zeng, Jia and others},
  booktitle={International Conference on Learning Representations},
  volume={2026},
  pages={60580--60606},
  year={2026}
}

@article{chen2026rmbench,
  title={Rmbench: Memory-dependent robotic manipulation benchmark with insights into policy design},
  author={Chen, Tianxing and Wang, Yuran and Li, Mingleyang and Qin, Yan and Shi, Hao and Li, Zixuan and Hu, Yifan and Zhang, Yingsheng and Wang, Kaixuan and Chen, Yue and others},
  journal={arXiv preprint arXiv:2603.01229},
  year={2026}
}

@inproceedings{chung2026rethinking,
  title={Rethinking progression of memory state in robotic manipulation: An object-centric perspective},
  author={Chung, Nhat and Hanyu, Taisei and Nguyen, Toan and Le, Huy and Bumgarner, Frederick and Nguyen, Duy Minh Ho and Vo, Khoa and Yamazaki, Kashu and Rainwater, Chase and Kieu, Tung and others},
  booktitle={Proceedings of the AAAI Conference on Artificial Intelligence},
  volume={40},
  number={5},
  pages={3407--3415},
  year={2026}
}

@article{lipman2022flow,
  title={Flow matching for generative modeling},
  author={Lipman, Yaron and Chen, Ricky TQ and Ben-Hamu, Heli and Nickel, Maximilian and Le, Matt},
  journal={arXiv preprint arXiv:2210.02747},
  year={2022}
}

@article{jiang2026ar,
  title={AR-WAM: a Visual-Conditioned Agent-Ready world action model for robotic manipulation},
  author={Jiang, Yicheng and Gan, Zesen and Wang, Xiaobo and He, Tianlun and Zhao, Chenxu and Wu, Minghui and Wang, Xinyue and Wang, Jiaxu and He, Junhao and Wang, Jianan and others},
  journal={arXiv preprint arXiv:2609.23578},
  year={2026}
}

@article{hu20262am,
  title={2AM: Grounding Agent-Side Memory as Guidance for Steerable Action Models in Long-Horizon Manipulation},
  author={Hu, Yutong and Chen, Fengjiao and Cao, Xuezhi and Detry, Renaud},
  journal={arXiv preprint arXiv:2609.11308},
  year={2026}
}

@article{jiang2026memoryvam,
  title={MemoryVAM: Integrating Memory into Video Action Model for Robot Manipulation},
  author={Jiang, Yuxin and Yu, Chang and Chen, Yunuo and Feng, Xiang and Yang, Yin and Gite, Nishank and Jiang, Chenfanfu},
  journal={arXiv preprint arXiv:2606.20679},
  year={2026}
}

@article{zhao2026remember,
  title={Remember What You Did: Action-History Memory with Dual-Expert Denoising for Long-Horizon Vision-Language-Action Policies},
  author={Zhao, Yaxin and Huang, Dianye and Wang, Chenwei and Yang, Chenguang and Jiang, Zhongliang},
  journal={arXiv preprint arXiv:2609.37307},
  year={2026}
}

@inproceedings{shi2026memoryvla,
  title={Memoryvla: Perceptual-cognitive memory in vision-language-action models for robotic manipulation},
  author={Shi, Hao and Xie, Bin and Liu, Yingfei and Sun, Lin and Liu, Fengrong and Wang, Tiancai and Zhou, Erjin and Fan, Haoqiang and Zhang, Xiangyu and Huang, Gao},
  booktitle={International Conference on Learning Representations},
  volume={2026},
  pages={18567--18602},
  year={2026}
}

@article{liu2023libero,
  title={Libero: Benchmarking knowledge transfer for lifelong robot learning},
  author={Liu, Bo and Zhu, Yifeng and Gao, Chongkai and Feng, Yihao and Liu, Qiang and Zhu, Yuke and Stone, Peter},
  journal={Advances in Neural Information Processing Systems},
  volume={36},
  pages={44776--44791},
  year={2023}
}

@article{maaten2008visualizing,
  title={Visualizing data using t-SNE},
  author={Maaten, Laurens van der and Hinton, Geoffrey},
  journal={Journal of machine learning research},
  volume={9},
  number={Nov},
  pages={2579--2605},
  year={2008}
}

@article{gu2026safe,
  title={Safe: Multitask failure detection for vision-language-action models},
  author={Gu, Qiao and Ju, Yuanliang and Sun, Shengxiang and Gilitschenski, Igor and Nishimura, Haruki and Itkina, Masha and Shkurti, Florian},
  journal={Advances in Neural Information Processing Systems},
  volume={38},
  pages={40041--40076},
  year={2026}
}

@article{chen2025robotwin,
  title={Robotwin 2.0: A scalable data generator and benchmark with strong domain randomization for robust bimanual robotic manipulation},
  author={Chen, Tianxing and Chen, Zanxin and Chen, Baijun and Cai, Zijian and Liu, Yibin and Li, Zixuan and Liang, Qiwei and Lin, Xianliang and Ge, Yiheng and Gu, Zhenyu and others},
  journal={arXiv preprint arXiv:2506.18088},
  year={2025}
}

@article{zhang2026unexpected,
  title={An unexpected robot policy: Early evaluations of GPT-6 Astra on RoboDojo and beyond},
  author={Zhang, Wenbo and Wang, Kaixuan and Ouyang, Yutao and Huang, Xiaoyu and Li, Liyang and Su, Kailun and Jin, Weiyang and Chai, Wenhao and Liang, Haotian and Dou, Zhiyang and others},
  journal={arXiv preprint arXiv:2609.24170},
  year={2026}
}

@article{liu2023reflect,
  title={Reflect: Summarizing robot experiences for failure explanation and correction},
  author={Liu, Zeyi and Bahety, Arpit and Song, Shuran},
  journal={arXiv preprint arXiv:2306.15724},
  year={2023}
}

@article{huang2022inner,
  title={Inner monologue: Embodied reasoning through planning with language models},
  author={Huang, Wenlong and Xia, Fei and Xiao, Ted and Chan, Harris and Liang, Jacky and Florence, Pete and Zeng, Andy and Tompson, Jonathan and Mordatch, Igor and Chebotar, Yevgen and others},
  journal={arXiv preprint arXiv:2207.05608},
  year={2022}
}

@article{sun2026himem,
  title={HiMem-WAM: Hierarchical Memory-Gated World Action Models for Robotic Manipulation},
  author={Sun, Xiaoquan and Zhang, Ruijian and Cao, Chen and Sun, Yihan and Chen, Jiahui and Xu, Zetian and Chen, Bo and Chen, Haijier and Yang, Zhen and Zhu, Jiarun and others},
  journal={arXiv preprint arXiv:2606.10363},
  year={2026}
}
\bibliographystyle{reference}
}

% \clearpage
\appendix
\clearpage

\section*{Appendix Contents}
\begingroup
% \footnotesize
\setlength{\parindent}{0pt}
\newcommand{\apptocline}[3][0em]{%
  \noindent\hspace*{#1}\hyperref[#2]{\ref*{#2}\hspace{0.6em}#3}%
  \nobreak\leaders\hbox{$\mkern2mu.\mkern2mu$}\hfill\nobreak\pageref*{#2}\par}
\newcommand{\appsec}[2]{\vspace{0.2em}\apptocline{#1}{\textbf{#2}}}
\newcommand{\appsub}[2]{\apptocline[1.6em]{#1}{#2}}
\newcommand{\appsubsub}[2]{\apptocline[3.2em]{#1}{#2}}

\appsec{app:setting}{Detailed Experimental Settings}
\appsub{app:data}{Datasets and Preprocessing}
\appsub{app:baseline}{Details of Baselines}
\appsub{app:implementation}{Details of Implementations}

\appsec{app:extended_exp}{Extended Experiments}
\appsub{app:runtime_efficiency}{End-to-End Runtime Efficiency}
\appsub{app:ablation_memory}{Ablation Study on Memory}
\appsub{app:adaptive_vlm}{Analysis of Adaptive VLM Invocation}

\appsec{app:limitations}{Limitations and Outlook}

\endgroup

\section{Detailed Experimental Settings}
\label{app:setting}

\subsection{Datasets and Preprocessing}
\label{app:data}

\textbf{RMBench.}
RMBench~\citep{chen2026rmbench} comprises nine bimanual manipulation tasks: five single-observation tasks, $M(1)$ (\textit{Observe and Pick Up}, \textit{Rearrange Blocks}, \textit{Put Back Block}, \textit{Swap Blocks}, and \textit{Swap T}), and four multi-observation tasks, $M(n)$ (\textit{Battery Try}, \textit{Blocks Ranking Try}, \textit{Cover Blocks}, and \textit{Press Button}). For WAM adaptation, we split the demonstrations of each task by episode at a 9:1 training-to-validation ratio, segmenting each episode into contiguous subtask intervals conditioned on the corresponding subtask instruction. We combine the head and wrist camera views as visual inputs and standardize joint states, gripper states, and actions using training-split statistics.

\textbf{LIBERO-Mem.}
LIBERO-Mem~\citep{chung2026rethinking} comprises ten single-arm manipulation tasks spanning three categories: two single-placement tasks, \texttt{P} (\textit{Bowl lift-and-return} and \textit{Bottle lift-and-return}), four counting tasks, \texttt{C} (\textit{Bowl lift-and-return $\times 3$}, \textit{Bottle lift-and-return $\times 3$}, \textit{Bowl lift-and-return $\times 5$}, and \textit{Bowl lift-and-return $\times 7$}), and four sequencing and spatial tasks, \texttt{S} (\textit{Swap two bowls}, \textit{Rotate three bowls}, \textit{Center occupied basket}, and \textit{Center empty basket}). Within each task, successful demonstrations are split by episode at a 9:1 training-to-validation ratio and partitioned into sequential manipulation stages. Each stage pairs the scene and wrist camera views with the global instruction, active subtask, and stage index $k/n$ to disambiguate repeated cycles.

\textbf{Real-world tasks.}
We collect demonstrations on a Franka robot arm equipped with a scene camera and a wrist camera across two multi-stage tasks: \textit{garbage sorting} (sequentially placing green, blue, and red blocks into color-matched bins) and \textit{bread toasting} (inserting bread slices into separate toaster slots, transferring them sequentially onto a plate stack, and pressing the red button). Complete trajectories are split by episode at a 9:1 training-to-validation ratio prior to extracting contiguous subtask segments. Each segment pairs the global instruction and active subtask with both camera views, 7-DoF joint positions, and gripper opening.

\subsection{Details of Baselines}
\label{app:baseline}

We compare \textbf{\method{}} against four families of representative methods aligned with each evaluation suite, including \textbf{\textit{(i)}} \textbf{\textit{VLA-based policies}}, \textbf{\textit{(ii)}} \textbf{\textit{WAM-based policies}}, \textbf{\textit{(iii)}} \textbf{\textit{Agentic frameworks}}, \textbf{\textit{(iv)}} \textbf{\textit{Other policy baselines}}.

\noindent \ding{224} \textbf{\textit{VLA-based policies}} include:
\begin{itemize}[leftmargin=1.2em, topsep=2pt, itemsep=1pt]
    \item \textbf{$\boldsymbol{\pi}_{0.5}$}~\citep{intelligence2025pi_} co-trains on diverse robot demonstrations and web data to jointly predict semantic subtasks and continuous actions from visual-language inputs.
    \item \textbf{X-VLA}~\citep{zheng2026x} introduces learnable source-specific soft prompts within a shared transformer backbone for scalable cross-embodiment action generation.
    \item \textbf{MemoryVLA}~\citep{shi2026memoryvla} retrieves visual details and semantic cues from perceptual-cognitive memory, fusing them with live observations for diffusion decoding.
    \item \textbf{ActMem-VLA}~\citep{zhao2026remember} encodes executed action history via Mamba to guide early denoising steps before refining action sequences with a frozen action expert.
\end{itemize}

\noindent \ding{224} \textbf{\textit{WAM-based policies}} include:
\begin{itemize}[leftmargin=1.2em, topsep=2pt, itemsep=1pt]
    \item \textbf{HiMem-WAM}~\citep{sun2026himem} learns hierarchical motion-skill latents and updates task memory at skill transitions.
    \item \textbf{AR-WAM}~\citep{jiang2026ar} conditions scene dynamics and action generation on target boxes and skill tokens.
    \item \textbf{MemoryVAM}~\citep{jiang2026memoryvam} compresses past visual observations into compact memory tokens that jointly condition video prediction and action decoding.
\end{itemize}

\noindent \ding{224} \textbf{\textit{Agentic frameworks}} include:
\begin{itemize}[leftmargin=1.2em, topsep=2pt, itemsep=1pt]
    \item \textbf{2AM}~\citep{hu20262am} employs a multimodal agent to track episode history and steer a low-level policy via subtask instructions and optional 2D spatial cues.
    \item \textbf{MaP-WAM}~\citep{map2026} uses completed subtask records to generate language and visual plans that condition next-stage action and progress prediction.
    \item \textbf{VoLoAgent}~\citep{volo2026} uses a VLM orchestrator to monitor execution and switch between learned policies and manipulation tools for error recovery, sharing identical WAM weights with \method{} in real-world tests.
\end{itemize}

\noindent \ding{224} \textbf{\textit{Other policy baselines}} include:
\begin{itemize}[leftmargin=1.2em, topsep=2pt, itemsep=1pt]
    \item \textbf{DP}~\citep{chi2025diffusion} generates action chunks via visual-conditioned denoising under receding-horizon control.
    \item \textbf{SlotSSM}~\citep{chung2026rethinking} tracks objects across frames with persistent object-centric slots and retains per-slot temporal histories via a state-space model for action prediction.
    \item \textbf{Mem-0}~\citep{chen2026rmbench} conditions actions on a subtask-start visual anchor and recent sliding-window frames.
\end{itemize}

\subsection{Details of Implementations}
\label{app:implementation}

\textbf{WAM adaptation.}
We initialize the RMBench and LIBERO-Mem executors from the RoboTwin2.0-IDM and LIBERO-IDM checkpoints of Long-WAM~\citep{huang2026longwam}, respectively. Stage~1 trains rank-32 LoRA adapters ($\alpha=32$) on the query, key, value, and output projections of the Video DiT self- and cross-attention layers for 10k steps at a peak learning rate of $5\times 10^{-5}$. Stage~2 freezes the adapted video branch and fine-tunes all Action DiT parameters for 30k steps at a peak learning rate of $10^{-5}$, while keeping the VAE, text, and proprioceptive encoders frozen throughout. Both stages use AdamW ($\beta_1=0.9$, $\beta_2=0.95$, weight decay $0.01$) with a global batch size of 32 across four 80\,GB NVIDIA H100 GPUs, BF16 mixed-precision compute, and FP32 master weights. Each stage starts an independent cosine schedule with 500 warmup steps (1,500 steps for LIBERO-Mem action fine-tuning). Evaluating the training sampling distributions yields fixed-horizon calibration factors (Eq.~(7) of the main paper) of $C=1.14$ on RMBench and $C=1.23$ on LIBERO-Mem.

\textbf{Gate training and calibration.}
We curate the distillation dataset by rolling out chunkwise teacher supervision across successful RMBench episodes to collect labeled chunk boundaries (\textsc{Skip} vs.\ \textsc{Call}), grouping episodes that share a task-seed pair or duplicate feature transitions into the same split to prevent leakage. At each boundary, the gate pairs the 1,024-dimensional action-token transition $\Delta f_t = f_t^+ - f_t^-$ with the 512-dimensional active subtask embedding $e_t$. Each block is standardized and $\ell_2$-normalized before concatenation and joint standardization via $\mathcal{N}(\cdot)$, fitting all statistics on unique training transitions. The routing gate $f_\phi$ is a single-hidden-layer MLP (128 ReLU units, 0.2 dropout) trained for 100 full-batch epochs with AdamW (learning rate $10^{-3}$, weight decay $0.01$) under Eq.~(6) of the main paper. On the validation split, we select the highest logit threshold $\eta$ retaining at least 97\% \textsc{Call} recall under both uniform and inverse-frequency weighting, yielding $\eta = -4.71$ ($b_t = \mathbb{I}[r_t \ge \eta]$).

\textbf{Closed-loop supervision.}
At each scheduled RMBench check, \texttt{GPT-6-Astra} receives the structured context $\mathcal{X}_t$ comprising initial and current RGB views from the head and both wrist cameras, up to five temporally ordered recent head-camera frames, the global goal $G$ and active plan $\mathcal{P}_t$, end-effector poses and gripper openings $q_t$, and the memory record $m_t$ (\texttt{refs}, \texttt{held}, and \texttt{last\_tool}). Episode initialization and post-intervention verification always enforce mandatory VLM checks ($b_t = 1$). For fixed-interval comparisons, the WAM retains chunk horizon $H=32$ while querying the VLM every chunk (32 steps) or every second chunk (64 steps).

\textbf{Real-world execution.}
On each physical Franka task, \method{} and VoLoAgent share identical task-specific WAM executor weights for controlled comparison. Conditioned on live scene and wrist RGB views, 7-DoF joint positions, gripper opening, and global plus subtask instructions, the executor predicts $H=32$ action steps specifying absolute 7-DoF joint targets and gripper commands.

\section{Extended Experiments}
\label{app:extended_exp}

\subsection{End-to-End Runtime Efficiency}
\label{app:runtime_efficiency}

\begin{figure}[h]
\centering
\includegraphics[width=0.8\columnwidth]{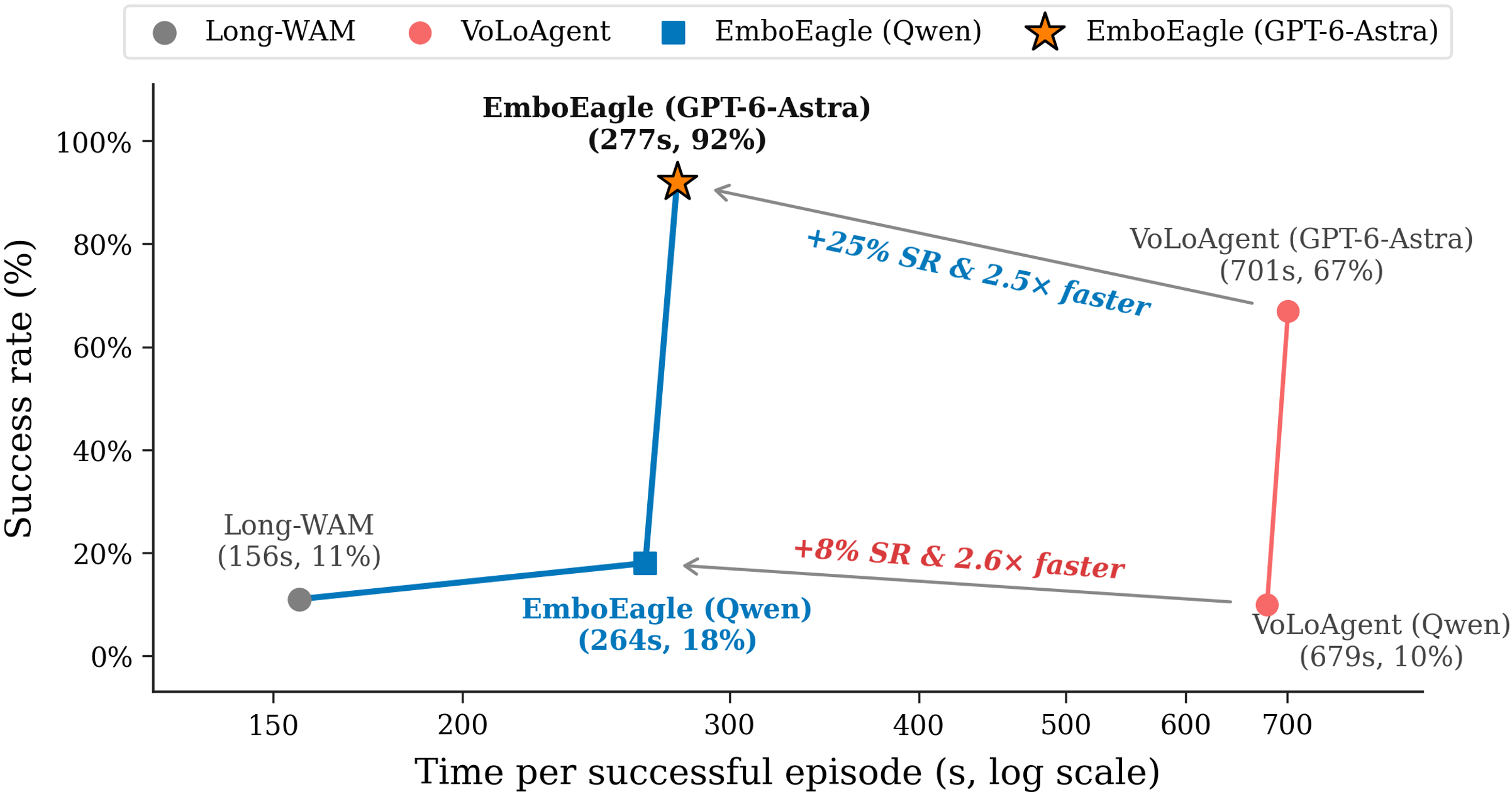}
% \vspace{-5pt}
\caption{\textbf{End-to-end accuracy--efficiency comparison on RMBench.} Task success rate (\%) versus mean completion time over successful episodes (s, log scale) across five configurations (10 test episodes per task), comparing open-loop Long-WAM, 64-step VoLoAgent, and adaptive \method{} under \texttt{Qwen3.6-27B} (thinking disabled) and \texttt{GPT-6-Astra}.}
\label{fig:app_eff}
% \vspace{-5pt}
\end{figure}

To evaluate how adaptive supervision translates into end-to-end execution speed, we compare task success rate against mean wall-clock time per successful episode across five configurations on RMBench (Figure~\ref{fig:app_eff}), sampling 10 test episodes per task. The vertical axis reports the proportion of successful test episodes, while the horizontal axis measures mean execution time (in seconds, log scale) computed exclusively over successful rollouts. As an open-loop baseline, Long-WAM~\citep{huang2026longwam} fine-tuned for 30k steps executes actions directly from the global task instruction, achieving 11\% success with an average duration of 156\,s. VoLoAgent~\citep{volo2026} adopts the 30k-step fine-tuned Long-WAM as its low-level executor and queries the VLM every two action chunks (64 control steps in total), whereas \method{} schedules VLM invocations adaptively during execution. Both frameworks are evaluated with \texttt{GPT-6-Astra} and \texttt{Qwen3.6-27B}, with thinking mode disabled across all \texttt{Qwen} configurations.

Across both VLM backbones, \method{} substantially improves task completion while cutting episode runtime by over 60\%. With \texttt{GPT-6-Astra}, \method{} raises success over VoLoAgent from 67\% to 92\% ($+25$ percentage points) while shortening mean successful episode time from 701\,s to 277\,s (a 60.5\% reduction, or $2.5\times$ faster). With \texttt{Qwen3.6-27B}, \method{} similarly lifts success from 10\% to 18\% ($+8$ percentage points) and reduces mean episode duration from 679\,s to 264\,s (a 61.1\% reduction, or $2.6\times$ faster).

\subsection{Ablation Study on Memory}
\label{app:ablation_memory}

To evaluate the contribution of structured recovery memory beyond VLM visual context, we ablate the structured memory \(m_t\) across four RMBench tasks (Table~\ref{tab:ablation_memory}). In \textit{w/o Structured Memory}, the VLM still observes the initial and recent frames in $\mathcal{V}_t$, but \texttt{assist} and \texttt{recovery} skills execute without cached initial poses (\texttt{refs}), active gripper--object bindings (\texttt{held}), or prior tool feedback (\texttt{last\_tool}). Removing $m_t$ reduces mean success across the four tasks from 97.0\% to 56.8\% ($-40.2$ points), demonstrating the importance of structured recovery memory beyond image-level history during physical repair.

Task-level sensitivity closely reflects geometric occlusion and pose-memory demands. The sharpest drop occurs on \textit{Swap T} ($100\% \to 21\%$), where each T-block must assume the other's initial 3D position and planar orientation; once the first block moves, its vacated pose is no longer recoverable from live RGB-D depth without \texttt{refs} and gripper-frame offsets in \texttt{held}. Similarly, removing memory costs 47 points on \textit{Cover Blocks} ($93\% \to 46\%$), where opaque lids occlude the colored blocks needed for ordered uncovering, and 32 points on \textit{Put Back Block} ($95\% \to 63\%$), where the block must return to its original mat after button pressing. By contrast, \textit{Rearrange Blocks} remains at 97\% without $m_t$ (vs.\ 100\% for \textbf{Ours}), as target mats and open gaps stay directly visible in live RGB-D observations for memory-free point primitives.

\begin{table}[t]
\centering
\caption{\textbf{Ablation study on structured memory.} Full-task success rates (\%) with and without structured memory $m_t$.}
\label{tab:ablation_memory}
% \vspace{-5pt}
\setlength{\tabcolsep}{16pt}
\renewcommand{\arraystretch}{1.2}
% \resizebox{0.90\columnwidth}{!}{
\begin{tabular}{l cccc}
\toprule
\textbf{Variants} & \textbf{Swap T} & \textbf{\makecell{Put Back\\Block}} & \textbf{\makecell{Rearrange\\Blocks}} & \textbf{\makecell{Cover\\Blocks}} \\
\midrule
w/o Structured Memory & 21\% & 63\% & 97\% & 46\% \\
\rowcolor{ThemeBlue!5}
\textbf{Ours} & \textbf{100\%} & \textbf{95\%} & \textbf{100\%} & \textbf{93\%} \\
\bottomrule
\end{tabular}
% }
% \vspace{-5pt}
\end{table}

\subsection{Analysis of Adaptive VLM Invocation}
\label{app:adaptive_vlm}

\textbf{Gate input ablation.}
We compare five input configurations for the routing gate across held-out episodes of all nine RMBench tasks (Table~\ref{tab:feature_ablation}), pairing the active subtask embedding $e_t$ with each candidate signal under the same MLP architecture and training protocol from Appendix~A.3. For each variant and split, the threshold $\eta$ is calibrated on validation episodes to achieve at least 97\% \textsc{Call} recall under both uniform and inverse-frequency weighting; test metrics use inverse-frequency weighting within each task and equal weight across tasks over three splits. Replacing external image features with the WAM's post-chunk action token $f_t^+$ raises average precision (AP) from 86.29\% to 91.92\% and F1 from 53.21\% to 67.02\% without an auxiliary vision encoder. Incorporating pre-execution context via the latent difference $\Delta f_t = f_t^+ - f_t^-$ further lifts AUROC to 98.04\%, AP to 92.62\%, and F1 to 73.58\% ($+6.56$ points over $f_t^+$ alone) while sharply reducing the cross-split F1 standard deviation from 9.74 to 1.32, retaining 98.15\% \textsc{Call} recall with a compact 197k-parameter gate.

\begin{table}[t]
\centering
\caption{\textbf{Gate input ablation on RMBench.} Routing performance (\%) when pairing subtask text $e_t$ with each candidate feature across three held-out splits (mean $\pm$ std).}
\label{tab:feature_ablation}
% \vspace{-5pt}
\setlength{\tabcolsep}{16pt}
\renewcommand{\arraystretch}{1.2}
% \resizebox{\columnwidth}{!}{
\begin{tabular}{l cccc}
\toprule
\textbf{\textsc{Feature}} & \textbf{\textsc{AUROC}} & \textbf{\textsc{AP}} & \textbf{\textsc{Recall of Call}} & \textbf{\textsc{F1}} \\  % \textsc
\midrule
$e_t$ only        & $66.52 \pm 2.36$ & $44.80 \pm 1.08$ & $97.37 \pm 2.42$ & $47.05 \pm 0.73$ \\
Image             & $92.49 \pm 1.38$ & $86.29 \pm 3.00$ & \underline{$98.56 \pm 1.87$} & $53.21 \pm 6.89$ \\
$f_t^-$           & $95.69 \pm 0.33$ & $90.99 \pm 1.80$ & $97.64 \pm 2.86$ & $66.24 \pm 13.33$ \\
$f_t^+$           & \underline{$96.61 \pm 1.53$} & \underline{$91.92 \pm 3.53$} & \bm{$98.92 \pm 0.94$} & \underline{$67.02 \pm 9.74$} \\
\rowcolor{ThemeBlue!5}
\textbf{$\Delta f_t$ (Ours)} & \bm{$98.04 \pm 1.03$} & \bm{$92.62 \pm 2.67$} & $98.15 \pm 0.78$ & \bm{$73.58 \pm 1.32$} \\
\bottomrule
\end{tabular}
% }
% \vspace{-5pt}
\end{table}

\textbf{Recall breakdown by teacher action.}
To examine which supervisory events are easiest or hardest to anticipate, we stratify teacher \textsc{Call} boundaries by top-level action $u_t \in \{\texttt{update}, \texttt{recovery}, \texttt{assist}\}$ using the deployed $\Delta f_t$ gate, averaging over the tasks in which each action occurs (Table~\ref{tab:gate_event_errors}). While overall \textsc{Call} recall reaches 98.15\%, stage transitions (\texttt{update}) and proactive assistance (\texttt{assist}) are captured almost without omission ($99.74\%$ and $100.00\%$ recall, with miss rates of $0.26\%$ and $0.00\%$). Reactive \texttt{recovery} exhibits a higher single-boundary miss rate of $7.91\%$ ($92.09\%$ recall), as incipient physical slips can initially induce subtle latent shifts; nevertheless, in closed-loop execution, persistent drift amplifies $\Delta f_t$ at subsequent chunk boundaries and still triggers timely repair.

\begin{table}[t]
\centering
\caption{\textbf{Gate recall breakdown by teacher action.} Per-action \textsc{Call} recall and miss rate (\%) of the $\Delta f_t$ gate averaged across tasks containing each event over three splits (mean $\pm$ std).}
\label{tab:gate_event_errors}
% \vspace{-5pt}
\setlength{\tabcolsep}{16pt}
\renewcommand{\arraystretch}{1.2}
\begin{tabular}{lccc}
\toprule
\textbf{Teacher Action} & \textbf{Tasks} & \textbf{\textsc{Call} Recall} & \textbf{Miss Rate} \\
\midrule
\texttt{update}   & 9 & $99.74 \pm 0.46$  & $0.26 \pm 0.46$ \\
\texttt{recovery} & 7 & $92.09 \pm 3.52$  & $7.91 \pm 3.52$ \\
\texttt{assist}   & 3 & $100.00 \pm 0.00$ & $0.00 \pm 0.00$ \\
\bottomrule
\end{tabular}
% \vspace{-5pt}
\end{table}

\section{Limitations and Outlook}
\label{app:limitations}

We note two main limitations of the current system. First, although our latent-delta routing gate eliminates 74.2\% of routine VLM checks, each triggered orchestration or post-intervention verification step still incurs non-trivial communication and inference latency when querying large VLM services, constraining deployment in strictly real-time or bandwidth-limited settings. Second, closed-loop reliability remains tied to the underlying VLM's embodied scene understanding and spatial reasoning capacity. While \method{} achieves substantial gains over prior orchestrators under a capable backbone, substituting smaller models degrades progress diagnosis and tool parameterization across multi-stage tasks. Together, these factors motivate post-training or distilling embodied diagnostic and spatial grounding capabilities from frontier supervisors into compact, on-device vision-language models, paired with asynchronous deliberation overlapped with low-level WAM execution to narrow the lightweight-backbone gap and advance real-time responsiveness.

\end{document}